\documentclass[10pt,journal,compsoc]{IEEEtran}

\usepackage{amssymb}
\usepackage{graphicx}
\usepackage[section]{algorithm}
\usepackage[noend]{algorithmic}
\usepackage{amsmath}
\usepackage{bm}
\usepackage{color}
\usepackage{ragged2e}
\usepackage{subfigure}
\usepackage[numbers,sort&compress]{natbib}
\usepackage{url}
\usepackage{enumitem}
\usepackage{fmtcount}

\usepackage{array}
\newcommand{\PreserveBackslash}[1]{\let\temp=\\#1\let\\=\temp}
\newcolumntype{C}[1]{>{\PreserveBackslash\centering}p{#1}}
\newcolumntype{R}[1]{>{\PreserveBackslash\raggedleft}p{#1}}
\newcolumntype{L}[1]{>{\PreserveBackslash\raggedright}p{#1}}

\renewcommand{\raggedright}{\leftskip=0pt \rightskip=0pt plus 0cm}

\usepackage{multirow}
\usepackage{longtable}
 \usepackage{rotating}

\begin{document}

\title{A Bi-layered Parallel Training Architecture for Large-scale Convolutional Neural Networks}

\author{Jianguo~Chen,
        Kenli~Li,~\IEEEmembership{Senior Member, IEEE},
        Kashif~Bilal,~\IEEEmembership{Member, IEEE},
        Xu Zhou,\\
        Keqin~Li,~\IEEEmembership{Fellow, IEEE},
        and Philip S. Yu,~\IEEEmembership{Fellow, IEEE}
\IEEEcompsocitemizethanks{\IEEEcompsocthanksitem Jianguo~Chen, Kenli~Li, Xu Zhou, and~Keqin~Li are with the College of Computer Science and Electronic Engineering, Hunan University, and the National Supercomputing Center in Changsha, Hunan, Changsha 410082, China.
\protect\\
Corresponding authors: Kenli Li (lkl@hnu.edu.cn) and Keqin Li (lik@newpaltz.edu).

\IEEEcompsocthanksitem Kashif Bilal is with COMSATS University Islamabad, Abbottabad 45550, Pakistan, and Qatar University, Doha 2713, Qatar.

\IEEEcompsocthanksitem Keqin Li is also with the Department of Computer Science, State University of New York, New Paltz, NY 12561, USA.

\IEEEcompsocthanksitem Philip S. Yu is with the Department of Computer Science, University of Illinois at Chicago, Chicago, IL 60607, USA, and Institute for Data Science, Tsinghua University, Beijing 100084, China.
}
}

\markboth{}
{Shell \MakeLowercase{\textit{et al.}}: Bare Advanced Demo of IEEEtran.cls for Journals}

\IEEEtitleabstractindextext{%
\begin{abstract}
 \raggedright{
Benefitting from large-scale training datasets and the complex training network, Convolutional Neural Networks (CNNs) are widely applied in various fields with high accuracy.
However, the training process of CNNs is very time-consuming, where large amounts of training samples and iterative operations are required to obtain high-quality weight parameters.
In this paper, we focus on the time-consuming training process of large-scale CNNs and propose a Bi-layered Parallel Training (BPT-CNN) architecture in distributed computing environments.
BPT-CNN consists of two main components:
(a) an outer-layer parallel training for multiple CNN subnetworks on separate data subsets, and (b) an inner-layer parallel training for each subnetwork.
In the outer-layer parallelism, we address critical issues of distributed and parallel computing, including data communication, synchronization, and workload balance.
A heterogeneous-aware Incremental Data Partitioning and Allocation (IDPA) strategy is proposed, where large-scale training datasets are partitioned and allocated to the computing nodes in batches according to their computing power.
To minimize the synchronization waiting during the global weight update process, an Asynchronous Global Weight Update (AGWU) strategy is proposed.
In the inner-layer parallelism, we further accelerate the training process for each CNN subnetwork on each computer, where computation steps of convolutional layer and the local weight training are parallelized based on task-parallelism.
We introduce task decomposition and scheduling strategies with the objectives of thread-level load balancing and minimum waiting time for critical paths.
Extensive experimental results indicate that the proposed BPT-CNN effectively improves the training performance of CNNs while maintaining the accuracy.
}
\end{abstract}

}

\maketitle
\IEEEdisplaynontitleabstractindextext
\IEEEpeerreviewmaketitle

\section{Introduction}
\label{intro}
\IEEEPARstart{I}{n} recent years, Deep Learning (DL) techniques have achieved promising results in various domains \cite{d01, d31}.
Convolutional Neural Network (CNN) algorithm is an important branch of DL.
Benefitting from large-scale training datasets and the complex training network, CNN achieves high accuracy and is widely applied in various domains, such as image classification \cite{d22}, speech recognition \cite{d16}, and text processing \cite{d09}.
However, the training process of CNN is very time-consuming, in which large amounts of training samples and iterative operations are required to obtain high-quality weight parameters \cite{d39, d19}.
It is critical to accelerate the training process and improve the performance of CNN.
Cloud computing, high-performance computing cluster, and supercomputing provides strong computing power for various applications \cite{d11, d12, d21}.
Therefore, it is a critical issue that how to design an effective parallel CNN training model based on distributed computing clusters and address the challenges of data communication, synchronization, and workload balancing, while maintaining high performance and high accuracy.

Numerous enhancements were proposed to accelerate the CNN and DL algorithms by improving the execution environments, mainly focusing in two directions \cite{d13, d24, d33, d11, d02, d17, d15}:
(1) using multi-core CPUs or GPUs platform to provide high-speed processing capacities,
and (2) using multiple distributed computers to increase the available computing power.
The CPU/GPU based methods \cite{d24, d33, d11} can perform more arithmetic operations and are suitable for the training of modestly sized DL models.
However, a known limitation of the these methods is that the training acceleration is small when the scale of training datasets or DL models exceeds the GPU memory capacity.
To process large-scale datasets and DL models, some distributed architecture based solutions were proposed, such as DistBelief\cite{d02}, Caffe \cite{d17}, and Tensorflow \cite{d15}.
Each of these approaches achieves significant progress in different perspectives, with implicit own complications in engineering and implementation.
Considerable improvements might be obtained by combining the advantages in both aspects and make use of the computing power of multiple machines in a distributed cluster and the high-performance CPUs or GPUs on each machine.

There exists multiple challenges in this regard.
Firstly, the entire CNN model contains multiple CNN subnetwork models that are trained in parallel on different machines, which requires synchronization and integration operations.
It is required to minimize the synchronization waiting problem between subnetwork models, and guarantee the accuracy of the integrated model.
Moreover, high-quality CNN models often require large-scale training dataset and a large number of iterations.
Hence, an effective parallel mechanism should be carefully designed to minimize the data communication overhead between different iteration steps, tasks in different threads/CPUs/GPUs, and distributed computers.
Furthermore, considering the heterogeneity of distributed computing clusters, computing nodes might be equipped with different CPU or GPU structures and have different training speed.
How to partition the training dataset into these computers and how many parallel training tasks are started on each computer to maximize the computing power and workload balance of each computer.
Finally, how to design a scalable parallel mechanism according to the characteristics of the CNN network, which can be easily deployed on elastic computing clusters and meet the application requirements in different areas.

In this paper, we aim to address the above challenges and fully utilize the parallel computing capacity of computing clusters and multi-core CPU to accelerate the training process of large-scale CNNs.
We propose a Bi-layered Parallel Training-CNN (BPT-CNN) architecture in distributed computing environments.
The outer-layer parallelism is deployed in a distributed cluster to train data subsets in parallel with minimal data communication and maximal workload balance.
The inner-layer parallelism is performed on each computer using multi-threaded platforms.
Experiments on large-scale datasets indicate the advantages of the proposed BPT-CNN in terms of performance, data communication, workload balance, and scalability.
The contributions of this paper are summarized as follows.

\begin{itemize}
    \item In the outer-layer parallelism, an Incremental Data Partitioning and Allocation (IDPA) strategy is proposed  to maximize the workload balance and minimize data communication among computers,
         where large-scale training datasets are partitioned and allocated to computers in batches according to their computing power.
    \item An Asynchronous Global Weight Updating (AGWU) strategy is proposed to integrate CNN subnetwork models from different computers and to address the synchronization waiting problem during the global weight update process.
    \item In the inner-layer parallelism, two time-consuming training steps of the CNN model are parallelized on each computer based on task-parallelism, including convolutional layer and local weight training process.
    \item To achieve thread-level load balancing and critical paths waiting time minimization, we introduce task decomposition and scheduling strategies for CNN training tasks with multi-threaded parallelism.
\end{itemize}

The rest of the paper is organized as follows.
Section \ref{section2} reviews the related work.
Section \ref{section3} presents the BPT-CNN architecture and the outer-layer parallelization process.
Section \ref{section5} describes the inner-layer parallel training of BPT-CNN.
Experimental results and evaluations are discussed in Section \ref{section6}.
Finally, Section \ref{section7} concludes the paper with a discussion of future work and research directions.

\section{Related Work}
\label{section2}
Previous works have proposed various hardware designs for CNNs and other deep learning algorithms acceleration \cite{d36,d11}.
FPGAs have been widely explored as hardware accelerators for CNNs because of their reconfigurability and
energy efficiency \cite{d04}.
In \cite{d13}, a parallel solution of CNNs was designed on many-core architecture, in which the model is parallelized on a new platform of Intel Xeon Phi Coprocessor with OpenMP.
Caffe \cite{d17} provides multimedia scientists and practitioners with a clean and modifiable framework for state-of-the-art deep learning algorithms.
Chung \emph{et al}. proposed a parallel Deep Neural Networks (DNNs) training approach for big data on the IBM Blue Gene/Q system, in which issues of regarding programming model and data-dependent imbalances were addressed \cite{d04}.
In \cite{d08}, a massively parallel coprocessor was designed as a meta-operator for CNNs, which consists of parallel 2D convolution primitives and programmable units.

To efficiently handle large-scale CNN and big data, outstanding distributed architecture based solutions were implemented in \cite{d37, d02}.
Adam \cite{d26} is an efficient and scalable deep learning training system, optimizing and balancing workload computation and communication through entire system co-design.
An energy-efficient reconfigurable accelerator was presented in \cite{d07} for deep CNN.
To minimize the energy cost of data movement for any CNN shape, a processing row stationary dataflow was introduced to reconfigure the computation mapping of a given shape.
In \cite{d02}, Dean \emph{et al}. introduced a distributed system (termed DistBelief) for training large neural networks on massive amounts of data.
DistBelief uses two complementary types of parallelism: distributed parallel between multiple models and in each model, respectively.
In addition, an asynchronous Stochastic Gradient Descent (SGD) procedure was employed to support a large number of model replicas.

Tensorflow \cite{d15} is a popular framework for large-scale machine learning on heterogeneous distributed systems.
The computation model of Tensorflow is based on dataflow graphs with mutable state, where the graph nodes can be distributed and executed in parallel on different workers, multi-core CPUs, and general-purpose GPUs.
In addition, Tensorflow uses a declarative programming paradigm, and developers can focus on the symbolic definition and computation logic instead of the implementation details.
However, there are some shortcomings in Tensorflow:
(a) TensorFlow attempts  to occupy all available GPU memory in the initial phase, which makes the machines deploying the Tensorflow program infeasible to share with other applications, and
(b) many high-level operations and interfaces in TensorFlow are nested and chaotically packaged, making it difficult to customize programming.
Hence, we study the parallelism idea of the DistBelief and Tensorflow approaches, and implement a bi-layered parallel training architecture for large-scale CNNs by combining the advantages of both distributed computing and CPU/GPU parallel computing.

Comparing with existing efforts, the proposed BPT-CNN architecture in this paper fully utilizes the parallel capacity of both the distributed cluster and multi-core CPU of individual machines.
Benefitting from the proposed IDPA and AGWU strategies, we effectively improve the training performance of the CNN model and address the problems of data communication, synchronization, and workload balancing of distributed cluster.
Moreover, according to task decomposition and scheduling strategies, BPT-CNN achieves the optimization objectives of thread-level load balancing and waiting time minimization of critical paths.

\section{BPT-CNN Architecture for CNNs}
\label{section3}

\subsection{Convolutional Neural Networks}
CNN model is one of the most representative network structures of DL technologies and has become one of the hot topics in various fields of science.
The common architecture of a CNN network includes two components: a feature extractor and a fully-connected classifier.
In a convolutional layer, each batch of the input dataset is analyzed to obtain different abstract features.
Given an input $X$ with scale ($D_{x}, H_{x}, W_{x}$), where $D_{x}$, $H_{x}$, and $W_{x}$ refer to the depth, height, and width of $X$.
Assuming that a filter with scale ($D_{f}, H_{f}, W_{f}$) is used in the convolutional layer to extract a feature map $A$, then $a_{i,j}$ denotes the value of the $j$-th column of the $i$-th row of the current feature map, as calculated in Eq. (\ref{eq01}):
{\small
\begin{equation}
\setlength{\abovecaptionskip}{0pt}
\setlength{\belowcaptionskip}{0pt}
\label{eq01}
a_{i,j} = f \left(\sum_{d=1}^{D_{f}}{\sum_{m=1}^{H_{f}}{\sum_{n=1}^{W_{f}}{(w_{d,m,n} \times x_{d,i,j})}}} + w_{b}\right),
\end{equation}
where} $D_{f}$, $H_{f}$, and $W_{f}$ are the depth, height, and width of the current filter, and $f()$ is an activation function, such as $sigmoid$, $tanh$, or $relu$ function \cite{d10}.

\begin{figure}[!ht]
 \setlength{\abovecaptionskip}{0pt}
 \setlength{\belowcaptionskip}{0pt}
 \centering
 \includegraphics[width=3.4in]{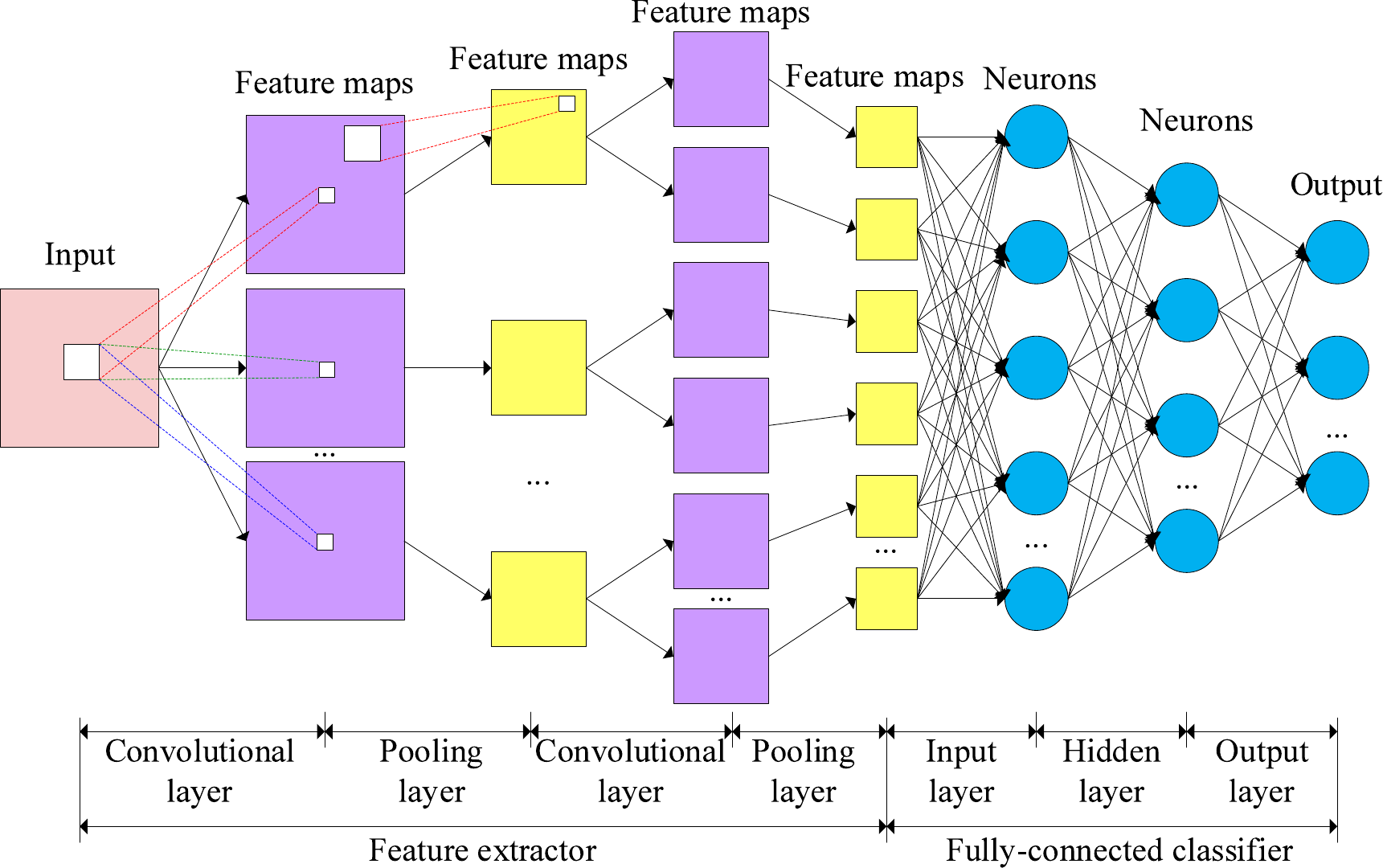}
 \caption{Example of a CNN architecture.}
 \label{fig01}
\end{figure}

Pooling layer is utilized on each feature map to reduce the feature dimensions.
There exist various pooling methods, i.e., max pooling and mean pooling.
Fully-connected layer is a classification layer of CNN, where all output features of convolutional or pooling layers are connected to all hidden neurons with weight parameters.
An example of a CNN architecture is illustrated in Fig. \ref{fig01}.

Massive training datasets and iterative training process guarantee the high precision of CNN.
However, they become the performance bottleneck when the training network structure is complex and the computing power is insufficient.
The characteristics and the corresponding limitations of CNNs are summarized as follows.

\begin{itemize}
    \item Massive weight parameters:
    There are a large number of weight parameters in the network.
    If there are different connection structures among layers, the network will become more complex, which requires more computation time cost.
    \item Large-scale training datasets and massive iterative operations for weight updating:
    High accuracy of CNNs requires high-quality weights, which depends on large-scale training datasets and iterative training.
\end{itemize}

\subsection{Bi-layered Parallel Training Architecture}
To accelerate the training process of CNNs, we propose a bi-layered parallel training architecture for large-scale CNNs.
We describe the distributed computing environment and training process of the BPT-CNN architecture.

\subsubsection{BPT-CNN Architecture}
\begin{figure}[!ht]
 \setlength{\abovecaptionskip}{0pt}
 \setlength{\belowcaptionskip}{0pt}
 \centering
 \includegraphics[width=3.4in]{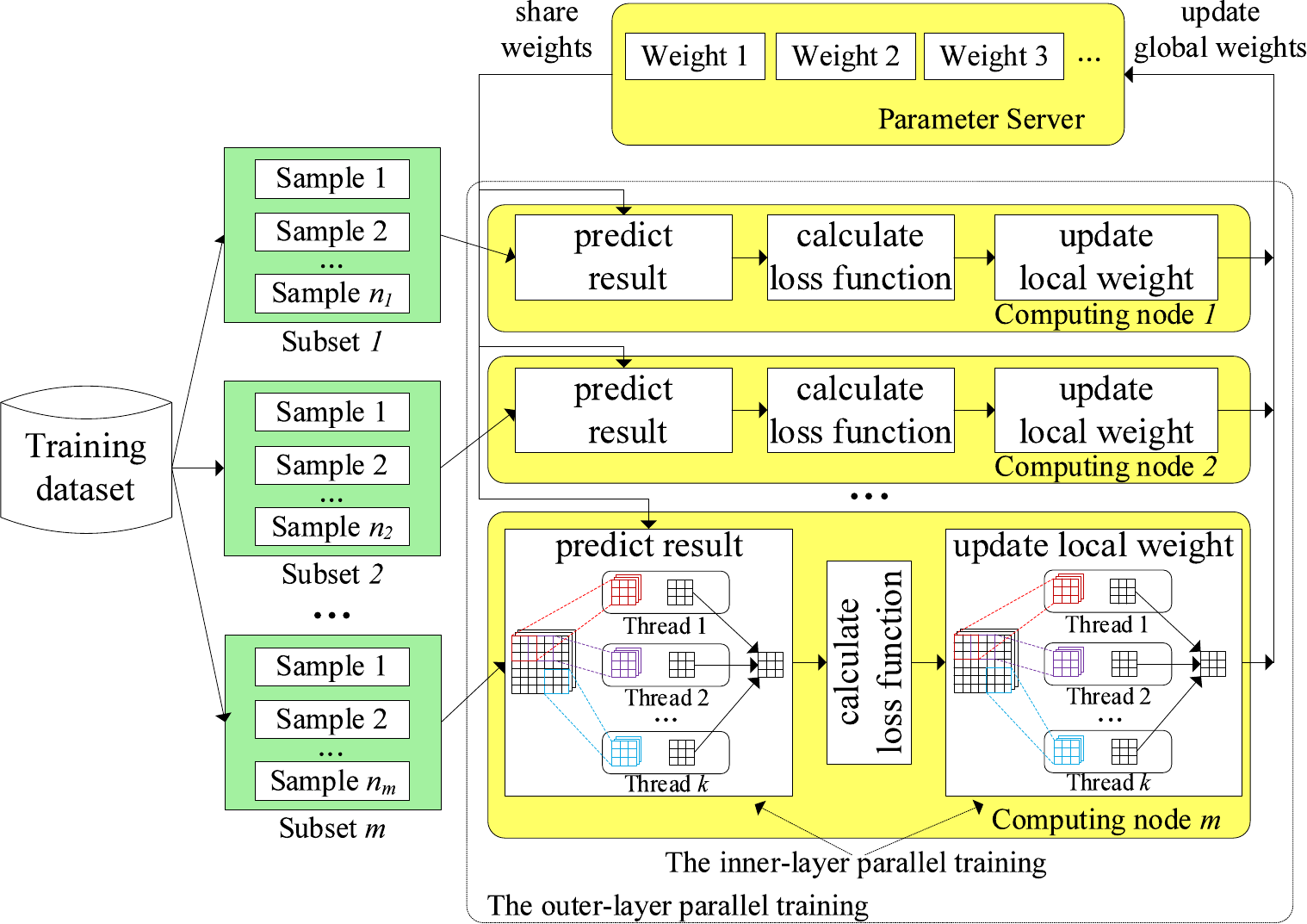}
 \caption{Bi-layered parallel training architecture for CNNs.}
 \label{fig02}
\end{figure}

BPT-CNN architecture is composed of two main components: (a) an outer-layer parallel training for multiple CNN subnetworks on separate data subsets, and (b) an inner-layer parallel training for each subnetwork.
The proposed BPT-CNN architecture is illustrated in Fig. \ref{fig02}.

(1) Outer-layer parallel training.
A data-parallelism strategy is adopted in the outer-layer parallel training process, where a large-scale dataset is split into multiple subsets and allocated to different computing nodes to be trained in parallel.
At the parameter server, the global weight parameters of the entire CNN network are updated depending on the local weights from each training branch.
The updated global weight parameters are shared to each machine for the next iterative training.

(2) Inner-layer parallel training.
The inner layer adopts a task-parallelism strategy to further accelerate the training process of each CNN subnetwork on each computer.
Two time-consuming computation tasks are parallelized, including convolutional layer and the local weight training process.
Computation tasks on these processes are decomposed depending upon their logical and data dependence, and are executed with multi-threaded parallelism.

\subsubsection{Distributed Computing Cluster for BPT-CNN}
\begin{figure}[!ht]
 \setlength{\abovecaptionskip}{0pt}
 \setlength{\belowcaptionskip}{0pt}
 \centering
 \includegraphics[width=3.45in]{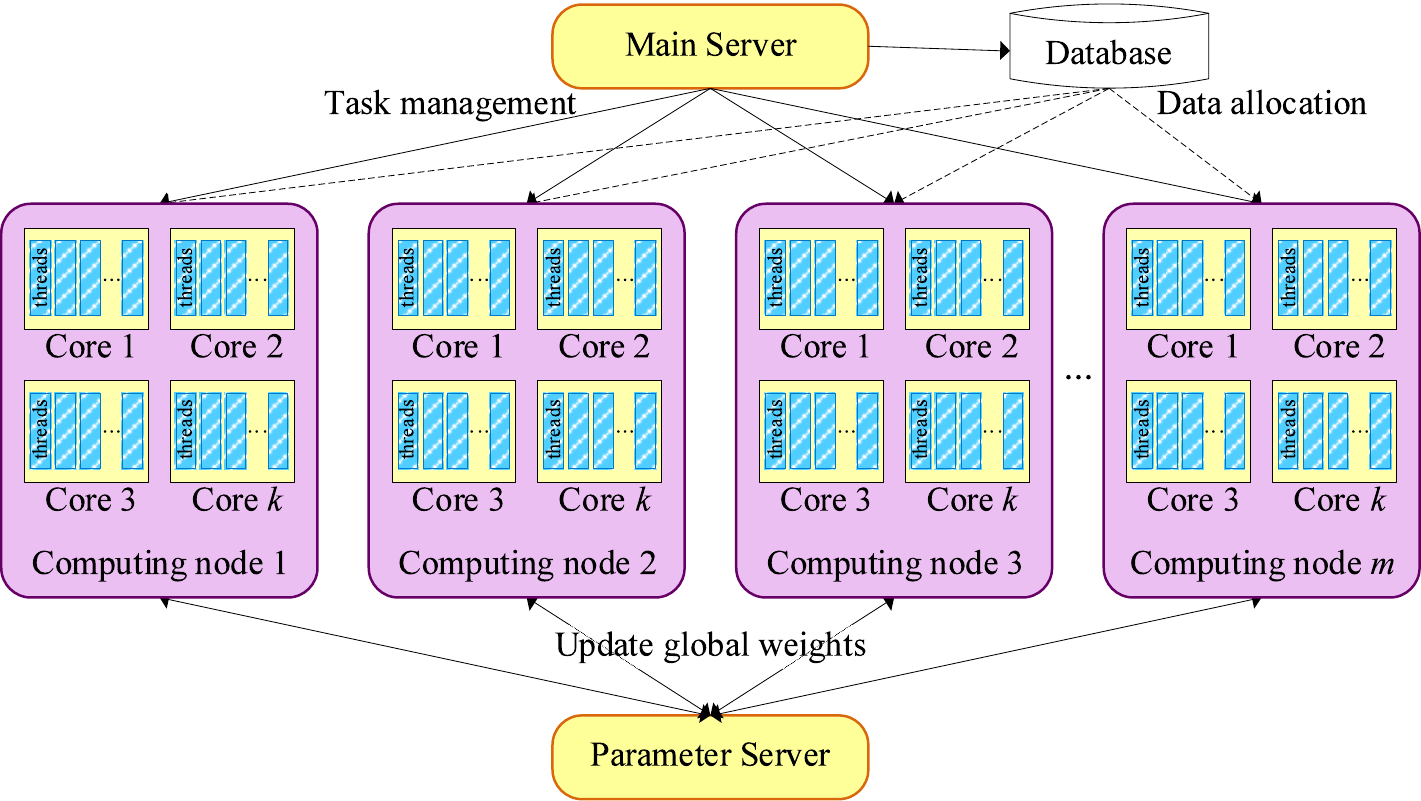}
 \caption{Structure of the distributed computing cluster for BPT-CNN.}
 \label{fig03}
\end{figure}

We construct a distributed computing cluster for the proposed BPT-CNN architecture to efficiently handle massive training datasets and large-scale CNN models.
The distributed cluster mainly consists of a main server, several computing nodes with mult-core CPU, and a parameter server, as shown in Fig. \ref{fig03}.

The main server is responsible for CNN training task management as well as data partition and allocation.
It copies the CNN training network and allocates them to each computing node.
Meanwhile, the training dataset is split into a series of subsets and allocated to the corresponding computing nodes.
During the parallel training, the main server monitors the training time costs on computing nodes and migrates datasets for the optimization objective of synchronization delay minimization.

On each computing node, samples in the subset are calculated by the corresponding CNN subnetwork, while the network weight parameters are trained as a local weight set.
The training process on each computer is executed in parallel.
In addition, each computer is equipped with a multi-core CPU platform.
In an inner-layer parallel training, the training process of each CNN subnetwork is further parallelized using multi-threaded parallelism.

The parameter server collects the trained local weight parameters from each computing node and updates the global weight parameters.
Then, the updated global weight set is re-allocated to each computing node for the next epoch of training.

\subsection{Outer-layer Parallel Training of BPT-CNN}
\label{section4}
In BPT-CNN's outer-layer parallel training architecture, we address critical issues of distributed and parallel computing, including data communication, synchronization, and workload balancing.
Firstly, considering the heterogeneity of the distributed computing cluster, we propose an incremental data partitioning and allocation strategy to maximize cluster's workload balancing and minimize data communication overhead.
In addition, we propose an asynchronous global weight updating strategy to further minimize the synchronous wait in the global weight update process.

\subsubsection{Incremental Data Partitioning and Allocation Strategy}
\label{section3.3.1}
Considering the heterogeneity of computing nodes and their different training speed, to maximize the workload balance of the distributed cluster and minimize the synchronization in global weight update process, we propose an Incremental Data Partitioning and Allocation (IDPA) strategy based on heterogeneous sensing.
As there are no dependencies between training samples, they can be partitioned and allocated in batches instead of done at once, according to the computing power of the computing nodes.
Assume that there are $N$ training samples, $m$ computing nodes in a distributed cluster, and $K$ training iterations are required for the CNN model.
Let $A$ ($A < K$) be the number of batches of data partitioning, that is, the entire training dataset is incrementally partitioned and allocated in $A$ times, and each time $\lfloor \frac{N}{A} \rfloor$ new samples are processed.

Initially, we take the first $\lfloor \frac{N}{A} \rfloor$ samples as the training dataset in the first batch.
Before the training iteration, we use the constant characteristics of the computing nodes to represent their heterogeneity, i.e., the CPU/GPU frequency is measured.
Let $\mu_{j}$ be the CPU/GPU frequency of computing node $C_{j}$, and the number of samples that will be partitioned and allocated to $C_{j}$ is calculated as:
\begin{equation}
\begin{aligned}
\label{eq02}
n_{j}^{(1)} &=
\left\{
\begin{array}{l l}
\left\lfloor  \lfloor \frac{N}{A} \rfloor \times \frac{\mu_{j}}{\sum\limits_{j'=1}^{m}{\mu_{j'}}}  \right\rfloor  & 1 \leq j \leq (m-1), \\
\lfloor \frac{N}{A} \rfloor - \sum\limits_{j'=1}^{m-1}{n_{j'}^{(1)}}  & j = m.\\
\end{array}
\right.
\end{aligned}
\end{equation}

After receiving the training samples, each computing node begins the first iteration of training.
At the same time, we monitor the execution time of each computing node to complete the iteration and evaluate its actual computing power.
Being of the opinion that there might be more applications from different employers executing on the compute nodes, although we can predict the computing power based on the computing node's CPU/GPU frequency, it is more accurate to evaluate its actual computing power by actual execution time.
Therefore, after the first training iteration, we can partition the training dataset according to the actual computing power of the machines.
Let $T_{j}$ be the execution time of computing node $C_{j}$ to train $n_{j}^{(1)}$ samples in the current iteration, then we can get the average execution time of $C_{j}$ for a sample as $\overline{t_{j}} = \frac{T_{j}}{n_{j}^{(1)}}$.
We collect the execution time of the computing nodes in the current iteration and predict the execution time required by all computing nodes in the next iteration.
Note that $\lfloor \frac{N}{A} \rfloor$ new samples will be partitioned and allocated in the $a$-th batch.
Namely, there is a total of $\lfloor \frac{N}{A} \rfloor \times a$ samples on the computing nodes.
The average execution time of the computing nodes in the $a$-th training iteration is calculated as:
\begin{equation}
\label{eq03}
T_{a} =  \frac{\lfloor \frac{N}{A} \rfloor \times a \times \overline{t}}{m},
\end{equation}
where $\overline{t} = \frac{1}{m}\sum_{j=1}^{m}{\overline{t_{j}}}$ is the average execution time for training a sample by any compute node.
To minimize the synchronization latency among computing nodes during the global weight update process, we expect all nodes to complete each iteration as close as possible.
Assume that $n'_{j}$ samples on $C_{j}$ after the $a$-th batch partitioning and allocation, we can calculate the value of $n'_{j}$ as:
\begin{equation}
\label{eq04}
n'_{j} =  \frac{T_{a}}{\overline{t_{j}}}.
\end{equation}
Accordingly, we can obtain the number of samples that $C_{j}$ can accept in the $a$-th batch allocation according to its actual computing power, as calculated as:
\begin{equation}
\label{eq05}
\begin{aligned}
n_{j}^{(a)} &=
\left\{
\begin{array}{l l}
n'_{j} - \sum\limits_{a'=1}^{a-1}n_{j}^{(a')} & 1 \leq j \leq m-1, \\
\lfloor \frac{N}{A} \rfloor - \sum\limits_{j'=1}^{m-1}{n_{j}^{(a)}}  & j = m.\\
\end{array}
\right.
\end{aligned}
\end{equation}

Repeat this process $A$ times until the entire training dataset is partitioned and allocated to the heterogeneous computing cluster. By considering the heterogeneity of computing nodes, each computing node receives a corresponding number of training samples based on its actual computing power.
The total number of samples allocated to each computing node $C_{j}$ is denoted as $n_{j}$, and $n_{j} = \sum_{a=1}^{A} {n_{j}^{(a)}}$.
The detail steps of the IDPA strategy are described in Algorithm \ref{alg01}.

\begin{algorithm}[!ht]
\scriptsize
\caption{{\scriptsize Incremental data partitioning and allocation strategy.}}
\label{alg01}
\begin{algorithmic}[1]
\REQUIRE ~\\
$N$: the number of samples in the training dataset;\\
$m$: the number of computing nodes in the distributed cluster;\\
$A$: the number of batches for data partitioning and allocation;\\
$a$: the current batch of data partitioning and allocation.\\
\ENSURE ~\\
$n_{s}$: the number list of sample partitioned to the computing nodes.\\
\IF {$a=1$}
\FOR {$j$ from 1 to $m-1$}
\STATE $n_{j}^{(1)} \leftarrow \left\lfloor \frac{\mu_{j}}{\sum_{j'=1}^{m}{\mu_{j'}}}  \times \lfloor \frac{N}{A} \rfloor \right\rfloor$; $n_{s}$.append($n_{j}^{(1)}$);
\ENDFOR
\STATE $n_{m}^{(1)} \leftarrow  \lfloor \frac{N}{A} \rfloor - \sum_{j'=1}^{m-1}{n_{j'}^{(1)}}$;  $n_{s}$.append($n_{m}^{(1)}$);
\ELSE
\STATE collect training duration $T_{j}$ from each computing node $C_{j}$ in the $(a-1)$-th iteration;
\STATE calculate the average training duration $\overline{t_{j}} \leftarrow \frac{T_{j}}{n_{j}^{(1)}}$ and $\overline{t} \leftarrow \frac{1}{m}\sum_{j=1}^{m}{\overline{t_{j}}}$;
\STATE predict the training duration of the $a$-th iteration $T_{a} \leftarrow  \frac{\lfloor \frac{N}{A} \rfloor \times a \times \overline{t}}{m}$;
\FOR {$j$ from 1 to $m-1$}
\STATE get the number of samples $n'_{j} \leftarrow  \frac{T_{a}}{\overline{t_{j}}}$ for $C_{j}$ in the $a$-th iteration;
\STATE $n_{j}^{(a)} \leftarrow n'_{j} - \sum_{a'=1}^{a-1}n_{j}^{(a')}$; $n_{s}$.append($n_{j}^{(a)}$);
\ENDFOR
\STATE $n_{m}^{(a)} \leftarrow  \lfloor \frac{N}{A} \rfloor - \sum_{j'=1}^{m-1}{n_{j}^{(a)}}$; $n_{s}$.append($n_{m}^{(a)}$);
\ENDIF
\RETURN $n_{s}$.
\end{algorithmic}
\end{algorithm}

Benefitting from the IDPA strategy, the training dataset is well partitioned and allocated to the computing nodes, allowing them to complete each iteration in same duration, achieving minimal synchronization delay and maximum workload balancing.
Moreover, no data migration is required among compute nodes during the training process, thereby no unnecessary data communication overhead is incurred.

Recall that $K$ iterations are required for the CNN model, that is, each sample has an average of $K$ times to train the weight parameter set of the CNN network model.
After the $A$ iterations in the data partitioning process, each computing node continues to execute the remaining iterations on the $n_{j}$ samples.
Since these samples are incrementally allocated to the computing nodes, the actual training times of samples in the $A$ iterations are $\frac{N}{A} \times \sum_{a=1}^{A}{a} = \frac{N(A+1)}{2}$ instead of $N \times A$.
Therefore, we should recalculate the remaining iterations $\Delta K$ of the training process, as defined below:
\begin{equation}
\label{eq06}
\begin{aligned}
\Delta K &=  \left\lfloor \frac{(N \times K) - \frac{N(A+1)}{2}}{N} \right\rfloor\\
         &= K -\frac{A}{2}-1.
\end{aligned}
\end{equation}
The total number of training iterations of the CNN model is $K' = A+\Delta K = K + \frac{A}{2}-1$.
To simplify the expression, we denote $K'$ as $K$ in the remaining context.

\subsubsection{Global Weights Updating Strategies}
There are massive connections with different weight parameters among all layers in a CNN network.
We define these weight parameters as a weight set.
We need to collect the training results on each computing node to update the global weight set for the entire CNN network.
In this section, we propose two global weight updating strategies for the CNN network.
We respectively define the local weight of each CNN subnetwork and the global weight of the entire CNN network as follows.

\textbf{Definition 1: Local weight set.}
\textit{
The weight parameters among all training layers of a CNN training network are denoted as a weight set.
At each computing node, the weight set of a CNN subnetwork is defined as the local weight set of the corresponding subnetwork.
The local weight set is trained based on the related data subset.
In a distributed computing cluster, there is a local weight set on each computing node, which is updated after training a sample.
}

\textbf{Definition 2: Global weight set.}
\textit{
The weight set of the entire CNN network is defined as the global weight set.
We provide a parameter server for calculating the global weight set by combining parts or all of the local weight sets.
The global weight set is aggregated by each local weight set and shared to all computing nodes for the next epoch of training.}

(1) Synchronous global weight updating strategy.

\begin{figure}[ht]
 \setlength{\abovecaptionskip}{0pt}
 \setlength{\belowcaptionskip}{0pt}
 \centering
 \includegraphics[width=3.4in]{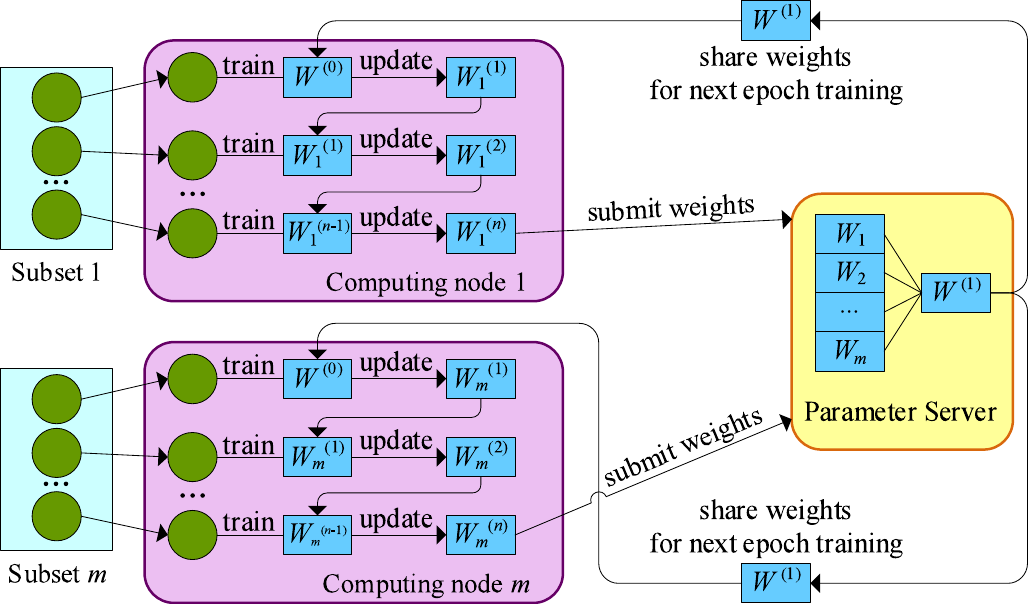}
 \caption{Synchronous global weight updating strategy. Initially, all computers use the same global weight set $W^{(0)}$ to train the first sample and get the corresponding outputs.
At each computer $C_{j}$, $W^{0}$ is updated to $W_{j}^{(1)}$ based on the first sample's output.
Then, $W_{j}^{(1)}$ is used to train the second sample and obtain $W_{j}^{(2)}$.
Repeat this step, until all samples on $C_{j}$ are trained, defining as an epoch of local iteration training.}
 \label{fig04}
\end{figure}

We propose a Synchronous Global Weight Updating (SGWU) strategy for BPT-CNN, where each computing node trains all the samples of the current subset and updates the local weight set for an iteration.
The local weight sets trained by all computing nodes in the current iteration are gathered at the parameter server, where a new version of the global weight set is generated.
The workflow of the SGWU strategy is illustrated in Fig. \ref{fig04}.

Considering that different local weight sets are trained by the corresponding subsets on different computers, having different contributions for the global weight set.
We verify the accuracy of each CNN subnetwork after completing an epoch of local iteration training and use it as the contribution of the local weight set.
After all computers finish an epoch of local iteration training, the latest local weight set trained on each computer is aggregated to the parameter server to update the global weight set as a new version $W^{(1)}$.
The global weight set $W^{(i)}$ for the $(i)$-th epoch of iteration training is defined in Eq. (\ref{eq07}):
{\small
\begin{equation}
\setlength{\abovecaptionskip}{0pt}
\setlength{\belowcaptionskip}{0pt}
\label{eq07}
W^{(i)} = \sum_{j=1}^{m}{W_{j}^{(i-1)} \times \frac{Q_{j}^{(i-1)}}{\sum_{k=1}^{m}{Q_{k}^{(i-1)}}}},
\end{equation}
where} $W_{j}^{(i-1)}$ and $Q_{j}^{(i-1)}$ are the local weight set and the corresponding accuracy of the CNN subnetwork on computer $C_{j}$, which is obtained in the $(i-1)$-th epoch of local iteration training.

In a distributed computing cluster,
especially for one equipped with heterogeneous computers, although we use the IDPA strategy to maximize cluster's workload balancing, the SGWU strategy inevitably faces the synchronization problem during the global weight update process.
Due to the different available computing capabilities, computers need different time costs to execute each training iteration.
Let $t_{j}^{(i)}$ be the execution duration for the $(i)$-th training iteration on computer $C_{j}$.
The waiting time for synchronization of the entire computing cluster is defined in Eq. (\ref{eq08}):
{\small
\begin{equation}
\setlength{\abovecaptionskip}{0pt}
\setlength{\belowcaptionskip}{0pt}
\label{eq08}
\mathbb{T}_{SGWU} = \sum_{i=1}^{K}{\sum_{j=1}^{m}{\left(\max_{j'=1}^{m}{(t_{j'}^{(i)})}-t_{j}^{(i)}\right)}},
\end{equation}
where} $K$ is the number of iteration training and $m$ is the number of computing nodes.

(2) Asynchronous global weight updating strategy.

\begin{figure}[!ht]
 \setlength{\abovecaptionskip}{0pt}
 \setlength{\belowcaptionskip}{0pt}
 \centering
 \includegraphics[width=3.4in]{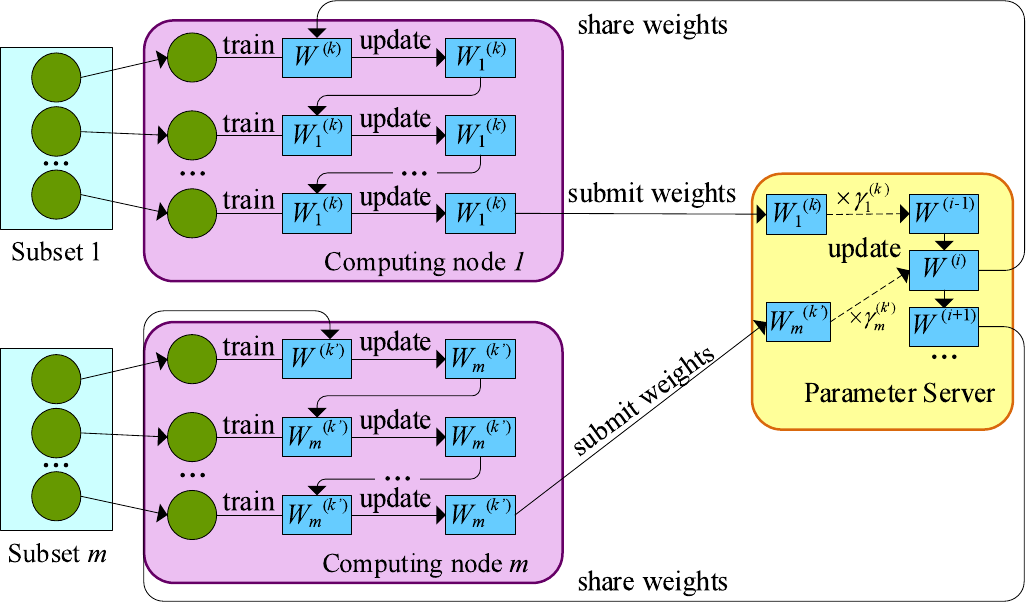}
 \caption{Asynchronous global weight updating strategy.
Each computer $C_{j}$ uses the current global weight set $W^{(k)}$ to train local samples and update the local weight set $W_{j}^{(k)}$.
Then, $W_{j}^{(k)}$ is submitted to the parameter server.
Note that the global weight set may have been updated from $W^{(k)}$ to $W^{(i)}$ by other computing nodes.
Now, based on the local weight set $W_{j}^{(k)}$, the global weight set $W^{(i)}$ is immediately updated to $W^{(i+1)}$.
Repeat this step, until all computing nodes complete the training iterations.}
 \label{fig05}
\end{figure}

To address the synchronization problem of SGWU, we propose an Asynchronous Global Weight Updating (AGWU) strategy.
In AGWU, once a computing node completes a training iteration on the local samples, the updated local weight set is submitted to the parameter server to immediately generate a new version of the global weight set, without waiting for other computing nodes.
Compared with SGWU, AGWU can effectively solve the synchronization waiting problem without increasing the communication overhead.
The workflow of the AGWU strategy is shown in Fig. \ref{fig05}.

Considering the heterogeneity of computing nodes, according to the IDPA strategy, each computing node may contain different scales of training subset.
In addition, due to the different training speeds, each computing node may also submit its local weight set to the parameter server at different time points, and get different versions of the global weight set.
For example, for a computing node $C_{j}$ with $n_{j}$ samples, we assume that $C_{j}$ train the local weight set $W_{j}^{(k)}$ based on based on the version $W^{k}$ of the global weight set in the current iteration.
During the current training iteration of $C_{j}$, the global weight set has been updated from $W^{(k)}$ to $W^{(i)}$ by other computing nodes.
In this case, the low speed computers train the local weight set based on the old version of the global weight set, while the high speed computers based on the newer version.
Denote ($W_{j}^{(k)} - W^{(k)}$) as the increment between the submitted local weight set $W_{j}^{(k)}$ and its base version global weight set $W^{(k)}$.
Assume that there is another local weight set $W_{j'}^{(i-1)}$ on $C_{j'}$ and it is trained based on the version $W^{(i-1)}$ of the global weight set, where $k < i-1$.
It is easy to know that $W_{j}^{(k)}$ has less impact than $W_{j}^{(k)}$ in the process of updating $W^{(i)}$.
Hence, we can conclude that the local weight sets using the old version of the global weight set have less impact on the global weight updating than those using the new version of the global weight set.
Therefore, we adopt a time attenuation factor $\gamma$ to measure the impact of each local weight set to the current global weight update process.
Denote $\gamma_{j}^{(k)}$ as the time attenuation factor of the local weight set $W_{j}^{(k)}$ submitted from $C_{j}$, as calculated in Eq. (\ref{eq09}):
{\small
\begin{equation}
\setlength{\abovecaptionskip}{0pt}
\setlength{\belowcaptionskip}{0pt}
\label{eq09}
\gamma_{j}^{(k)} = \frac{e^{\frac{k}{i-1}}}{\sum\limits_{\forall W_{j'}^{k'}, j' \neq j}{e^{\frac{k'}{i-1}}}},
\end{equation}
where} $(i-1)$ is the latest version of the global weight set, and $k$ is the version of the global weight set that used to train $W_{j}^{(k)}$.

Since there is no dependence among the training subsets on different computing nodes, the global weight update process does not require the training results from all computing nodes at the same time.
Once a local weight set is submitted, the current global weight set $W^{(i-1)}$ is immediately updated to a new version $W^{(i)}$, without waiting for other computing nodes.
The $i$-th version of the global weight set is updated as:
\begin{equation}
\setlength{\abovecaptionskip}{0pt}
\setlength{\belowcaptionskip}{0pt}
\label{eq10}
\begin{aligned}
W^{(i)} &= W^{(i-1)} + \Delta W_{j}^{k \rightarrow i}\\
        &= W^{(i-1)} + \gamma_{j}^{(k)} \times Q_{j}^{(k)}  \times (W_{j}^{(k)}- W^{(k)}), \\
\end{aligned}
\end{equation}
where $\Delta W_{j}^{k \rightarrow i}$ is the update component from $W_{j}^{(k)}$, and $Q_{j}^{(k)}$ is the accuracy of the CNN subnetwork on computer $C_{j}$, which is evaluated by the output of the current local iteration training.

After obtaining the updated global weight set $W^{(i)}$, the parameter server shares $W^{(i)}$ to $W_{j}$ for the next iteration training.
Subsequently submitted local weight sets from other computing nodes will update the global weight set based on the latest version.
The steps of the AGWU strategy of BPT-CNN is described in Algorithm \ref{alg02}.

\begin{algorithm}[!ht]
\scriptsize
\caption{{\scriptsize Asynchronous global weight updating strategy.}}
\label{alg02}
\begin{algorithmic}[1]
\REQUIRE ~\\
$W_{j}^{(k)}$: the local weight set from $C_{j}$ trained based on $W^{(k)}$;\\
$Q_{j}^{(k)}$: the accuracy of the CNN subnetwork model on $C_{j}$ in the $(k)$-th local iteration training.\\
\ENSURE ~\\
$W^{(i)}$: the new version of the global weight set.\\
\FOR {$j'$ from 1 to $m$}
\STATE find the version of the global weight set $W^{(k')}$ used for $W_{j'}^{(k')}$;
\STATE calculate time attenuation factor $\gamma_{j}^{(k)} = e^{\frac{k}{i-1}}/\sum_{\forall W_{j'}^{k'}, j' \neq j}{e^{\frac{k'}{i-1}}}$;
\ENDFOR
\STATE calculate the update component $\Delta W_{j}^{k \rightarrow i} \leftarrow \gamma_{j}^{(k)} \times Q_{j}^{(k)}  \times (W_{j}^{(k)}- W^{(k)})$;
\STATE update to the global weight set $W^{(i)} \leftarrow W^{(i-1)} + \Delta W_{j}^{k \rightarrow i} $;
\RETURN $W^{(i)}$.
\end{algorithmic}
\end{algorithm}

In comparison with the SGWU strategy, in the AGWU strategy, each computing node independently participates in the global weight update process, so there is no synchronization waiting problem in AGWU.
Furthermore, from the perspective of the entire training process, the update of the global weight set depends on the training results of all compute nodes.
According to Eq. (\ref{eq07}) and Eq. (\ref{eq10}), the global weight set is updated based on the local weight set and the corresponding accuracy of the trained mode in both of SGWU and AGWU strategies.

(3) Data communication of global weight updating.

In BPT-CNN, data communication only incurs between each computing node and the parameter server for the global weight updating and sharing.
In AGWU, to reduce the synchronization cost and data communication overhead from the perspective of computing nodes, after receiving a version of the global weight set, each computing node begin a training iteration.
It does not participate in the global weight update and receive a new version before completing the current iteration.
In both of SGWU and AGWU strategies, the global weight set is updated for every epoch of iteration training.
Therefore, both strategies produce the same data communication overhead.
Denote $K$ as the number of CNN iteration training, data communication $\mathbb{C}_{SGWU}$ in SGWU and $\mathbb{C}_{SGWU}$ in AGWU between the parameter server and all computing nodes is calculated in Eq. (\ref{eq11}):
{\small
\begin{equation}
\setlength{\abovecaptionskip}{0pt}
\setlength{\belowcaptionskip}{0pt}
\label{eq11}
\mathbb{C}_{AGWU} = \mathbb{C}_{SGWU} = 2c_{w} \times m \times K,
\end{equation}
where}
$c_{w}$ is a unit communication cost for transmitting the global weight set between the parameter server and a computing node.
For each update of the global weight set, there exist 2 iterations of data communication: (1) submitting the local weight set from a computing node to the parameter server, and (2) sharing the updated global weight set from the latter to the former.

\section{{\small Inner-layer Parallel Training of BPT-CNN}}
\label{section5}
In the inner-layer parallel training of BPT-CNN, we further parallelize the training process for each CNN subnetwork on each computing node.
Two time-consuming training steps are parallelized based on task-parallelism, including convolutional layer and the weight training process.
In addition, we propose task decomposition and scheduling solutions to realize thread-level load balancing and critical paths waiting time minimization.

\subsection{{\small Parallel Computing Models of CNN Training Process}}
\subsubsection{Parallelization of Convolutional Layer}
\label{section4.1.1}
In the training process of a CNN network, convolutional layers take more than 85.18\% of the total training duration, but only train 5.32 - 6.63\% of the weight parameters \cite{d14}.
Fortunately, the matrix-parallel-based method provides an effective way of performing convolutional operations in parallel.
We introduce the parallel mechanism of the convolutional operations into the inner-layer parallel training architecture of BPT-CNN.
We use the data partitioning method of the input matrix in CNN and extract all convolution areas from the input matrix.
Then, by sharing the filter matrix, all convolution areas are convoluted in parallel with the shared filter matrix.

Given an input matrix $X$ with the shape of ($D_{x} \times H_{x} \times W_{x}$), where $D_{x}$, $H_{x}$, and $W_{x}$ are the depth, height, and width of $X$.
Providing a filter parameter matrix $F$ with the shape of ($D_{f} \times H_{f} \times W_{f}$), a feature map $A$ is generated via convolutional multiplication on $X$ and $F$.
Based on the scales of $X$ and $F$, the shape of $A$ is calculated as:

{\small
\begin{equation}
\renewcommand{\arraystretch}{1.0}
\setlength{\abovecaptionskip}{0pt}
\setlength{\belowcaptionskip}{0pt}
\label{eq11}
\begin{aligned}
D_{a} &= D_{x} = D_{f},\\
H_{a} &= \frac{H_{x} - H_{f} + 2P}{S} +1,\\
W_{a} &= \frac{W_{x} - W_{f} + 2P}{S} +1,
\end{aligned}
\end{equation}
where} $D_{a}$, $H_{a}$, and $W_{a}$ are the depth, height, and width of $A$, respectively.
Based on the scales of $X$, $F$, and $A$, we calculate the times $K_{C}$ of convolutional operations in the current convolutional layer, which will be executed in parallel.
$K_{C}$ is calculated in Eq. (\ref{eq12}):

\begin{equation}
\footnotesize
\renewcommand{\arraystretch}{1.0}
\setlength{\abovecaptionskip}{0pt}
\setlength{\belowcaptionskip}{0pt}
\label{eq12}
\begin{aligned}
K_{C} &= \left(\frac{H_{x} - H_{f} + 2P}{S} +1 \right) \times \left(\frac{W_{x} - W_{f} + 2P}{S} +1\right),
\end{aligned}
\end{equation}
where $S$ is the stride of the convolutional operation and $P$ is the number of the zero padding, which means appending $P$ laps elements around $X$ with the value of 0.

To execute these $K_{C}$ operations in parallel, we need to identify the convolution areas of the input matrix $X$ for each task.
A convolution area $X[r_{begin} : r_{end}, ~c_{begin}: c_{end}]$ of $X$ includes the begin and end rows and columns.
In each convolutional operation task, an element-by-element multiplication is executed on $X[r_{begin} : r_{end}, ~c_{begin}: c_{end}]$ and $F$ to generate the corresponding element $a_{i,j}$ of $A$.
For each element $a_{i,j}$ in $A$, location indexes of the convolution area in $X$ is calculated in Eq. (\ref{eq13}):
{\small
\begin{equation}
\setlength{\abovecaptionskip}{0pt}
\setlength{\belowcaptionskip}{0pt}
\label{eq13}
\begin{aligned}
r_{begin} &= i \times S, ~~~r_{end} = r_{begin} + H_{f},\\
c_{begin} &= j \times S, ~~~c_{end} = c_{begin} + W_{f}.
\end{aligned}
\end{equation}}

\begin{figure}[!ht]
 \setlength{\abovecaptionskip}{0pt}
 \setlength{\belowcaptionskip}{0pt}
 \centering
 \includegraphics[width=3.45in]{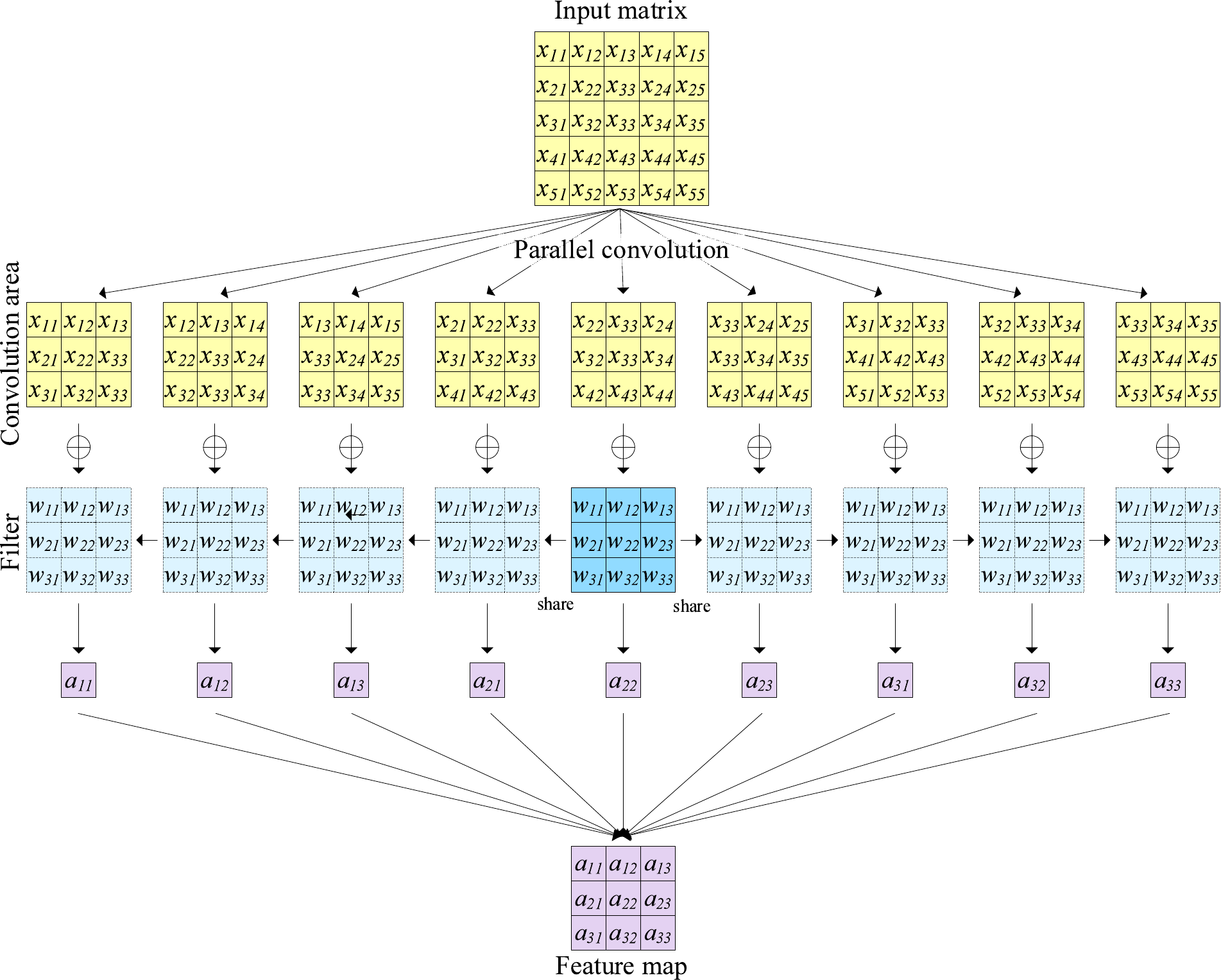}
 \caption{Parallel convolutional operation based on task-parallelism.}
 \label{fig06}
\end{figure}

After obtaining location indexes of each convolution area, we extract the contents of different convolution areas and perform the related convolutional operations in parallel, without waiting for the end of the previous convolutional operations.
These parallel convolutional operations on different areas access the input and filter matrices repeatedly and simultaneously from the same memory without updating the contents.
Without data dependence among these tasks, different tasks can access different convolution areas in $X$ simultaneously.
An example of the parallel convolutional operation of each CNN subnetwork in BPT-CNN is illustrated in Fig. \ref{fig06} and the steps of this process are described in Algorithm \ref{alg04}.

\begin{algorithm}[!ht]
\scriptsize
\caption{{\scriptsize Parallel convolutional operation of BPT-CNN.}}
\label{alg04}
\begin{algorithmic}[1]
\REQUIRE ~\\
$X$: The input training dataset;\\
$F$: the filter parameter matrix.\\
\ENSURE ~\\
$PT_{Conv}$: the parallel subtasks of the current convolutional layer.\\
\STATE calculate the size ($D_{a}$, $H_{a}$, $W_{a}$) of feature map $A$ in Eq. (\ref{eq11});
\STATE calculate convolution operation times $K_{C}$ in Eq. (\ref{eq12});
\FOR {each $k$ in $K_{C}$}
\STATE  get convolution area $X[r_{begin} : r_{end}, ~c_{begin}: c_{end}]$;
\STATE generate subtask $T_{k}$ $\leftarrow$  $Conv(X[r_{begin} : r_{end}, ~c_{begin}: c_{end}], F, a_{i,j})$;
\STATE  append to the parallel task list $PT_{Conv}$ $\leftarrow$ $T_{k}$;
\ENDFOR
\RETURN $PT_{Conv}$.
\end{algorithmic}
\end{algorithm}

As defined in Eq. (\ref{eq12}), the maximum parallelism degree of a convolutional layer is equal to the number of elements of the output feature map, which is computed according to the scale of $X$ and $F$.
The total execution duration $\mathbb{T}_{Conv}$ of a convolutional layer is calculated in Eq. (\ref{eq14}):
{\small
\begin{equation}
\setlength{\abovecaptionskip}{0pt}
\setlength{\belowcaptionskip}{0pt}
\label{eq14}
\mathbb{T}_{Conv} = \max_{i=1}^{|A|}{\mathbb{T}_{i}},
\end{equation}
where} $|A| = H_{A} \times W_{A}$ is the number of elements in $A$ and $\mathbb{T}_{i}$ is the execution duration of the $i$-th operation task.

\subsubsection{Parallelization of Local Weight Training Process}
To distinguish the weight set of the entire CNN network and that of each CNN subnetwork, we respectively define the global weight set and local weight sets in Section 3.3.2.
In this section, training process of the local weight set of each CNN subnetwork is parallelized on each computer.

After obtaining the outputs of a CNN subnetwork, the error (loss function) of each layer is evaluated from the output layer to the first convolutional layer using the Back Propagation (BP) method.
The Stochastic Gradient Descent (SGD) process \cite{d34, d32} is involved in updating the weight parameters among all layers of the current CNN subnetwork.
In the output layer, the square error of all neurons is taken as the objective function of weight training, as defined in Eq. (\ref{eq18}):
{\small
\begin{equation}
\setlength{\abovecaptionskip}{0pt}
\setlength{\belowcaptionskip}{0pt}
\label{eq18}
E_{x} = \sum_{i \in L_{output}}{(y_{i}' - y_{i})^{2}},
\end{equation}
where} $E_{x}$ denotes the loss function of the input $x$, and $y_{i}'$ and $y_{i}$ are the label and the output of the neuron $a_{i}$ in the output layer, respectively.
The error $\delta_{i}$ of $a_{i}$ is the inverse of the partial derivative of the error of the input of $a_{i}$, as calculated in Eq. (\ref{eq19}):

{\small
\begin{equation}
\setlength{\abovecaptionskip}{0pt}
\setlength{\belowcaptionskip}{0pt}
\label{eq19}
\begin{aligned}
\delta_{i} = -\frac{\partial E_{x}}{\partial net_{i}}
           =-\frac{\partial E_{x}}{\partial \sum_{j}{w_{ji}x_{ji}}},
\end{aligned}
\end{equation}
where} $x_{ji}$ is the input of the neuron $a_{i}$ that connected with $a_{j}$, that is, $x_{ji}$ is the output of $a_{j}$.
$w_{ji}$ is the weight of the connection between neurons $a_{j}$ and $a_{i}$.

Let $\delta^{l}$ be the set of errors of neurons in the $l$-th layer $L_{l}$.
Based on $\delta^{l}$, the error set $\delta^{l-1}$ of neurons in $L_{l-1}$ is calculated in Eq. (\ref{eq20}):
{\small
\begin{equation}
\setlength{\abovecaptionskip}{0pt}
\setlength{\belowcaptionskip}{0pt}
\label{eq20}
\delta^{l-1} = \sum_{i=1}^{N}{\delta_{i}^{l} \times W_{i}^{l-1} \oplus f^{'}(net^{l-1})},
\end{equation}
where} $W^{l-1}$ is the weight set of $L_{l-1}$ and $net^{l-1}$ is the weighted input of $L_{l-1}$, as defined as:
{\small
\begin{equation}
\setlength{\abovecaptionskip}{0pt}
\setlength{\belowcaptionskip}{0pt}
\label{eq21}
\begin{aligned}
net^{l-1} &= conv(W^{l-1}, a^{l-2}) + w_{b},\\
a_{i,j}^{l-2} &= f^{l-2}(net_{i,j}^{l-2}),
\end{aligned}
\end{equation}
where} $a^{l-2}$ is the output matrix of $L_{l-2}$, consisting of each element $a_{i,j}^{l-2}$.
An example of the calculation process of loss function between layers $L_{l}$ and $L_{l-1}$ is shown in Fig. \ref{fig07}.

\begin{figure}[!ht]
 \setlength{\abovecaptionskip}{0pt}
 \setlength{\belowcaptionskip}{0pt}
 \centering
 \includegraphics[width=2.6in]{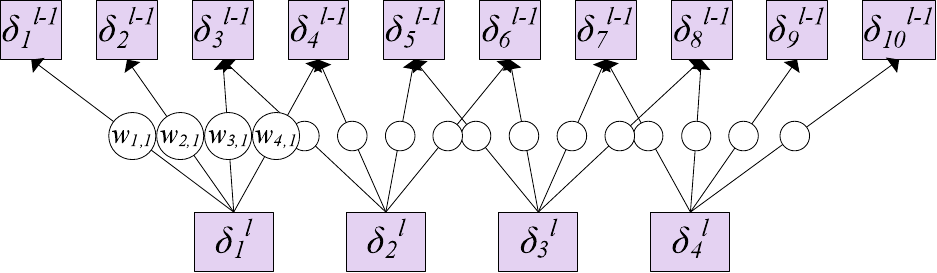}
 \caption{Example of the loss function computation.}
 \label{fig07}
\end{figure}

\begin{figure}[!ht]
 \setlength{\abovecaptionskip}{0pt}
 \setlength{\belowcaptionskip}{0pt}
 \centering
 \includegraphics[width=3.0in]{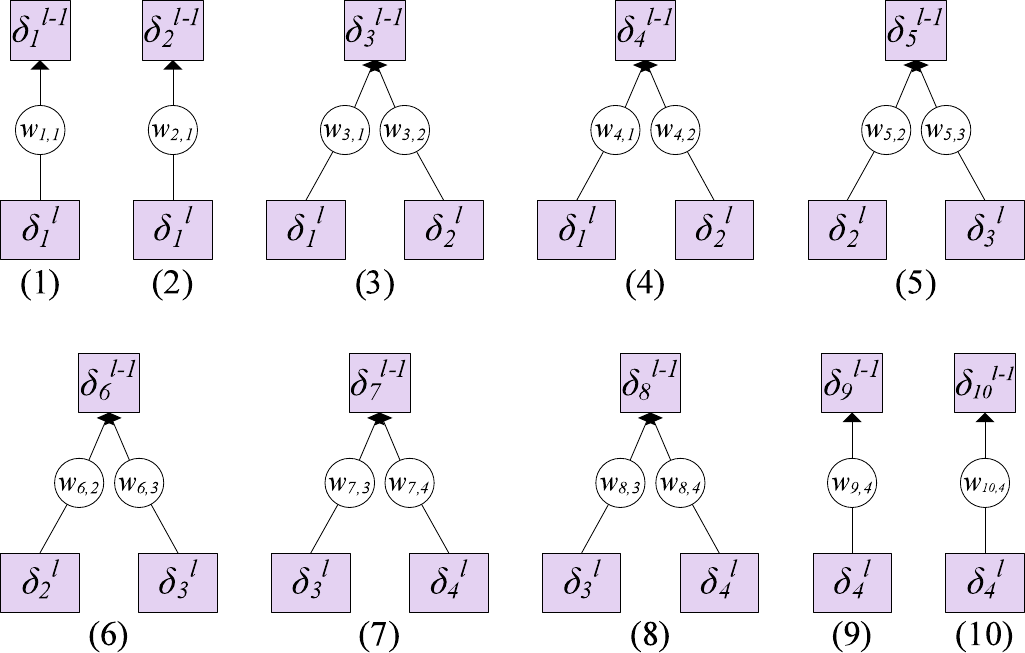}
 \caption{Example of the loss function parallel computation.}
 \label{fig08}
\end{figure}

We parallelize the process for the loss function calculation, where the errors of {neurons in the same layer are} computed in parallel.
In the convolutional layer, each neuron in the output layer (a feature map) is connected to a part of neurons in the input layer (an input matrix).
In such a case, the error calculation of neurons in the previous layer $L_{l-1}$ depends on the results of a part of neurons in the next layer $L_{l}$.
Hence, we parallelize this process depending on neurons in $L_{l-1}$.
An example of the loss function calculation parallelization is shown in Fig. \ref{fig08}.

After obtaining the error set of neurons in $L_{l}$, we calculate the error of each neuron in $L_{l-1}$.
Let $\delta_{i,j}^{l-1}$ be the error component of neuron $a_{j}$ in $L_{l}$ for $a_{i}$ in $L_{l-1}$, as defined as:
{\small
\begin{equation}
\setlength{\abovecaptionskip}{0pt}
\setlength{\belowcaptionskip}{0pt}
\label{eq22}
\begin{aligned}
\delta_{i,j}^{l-1} = \frac{\partial E_{d}}{\partial net_{i,j}^{l-1}}
=\frac{\partial E_{d}}{\partial a_{i,j}^{l-1}} \frac{\partial a_{i,j}^{l-1}}{\partial net_{i,j}^{l-1}},
\end{aligned}
\end{equation}
where} $H_{f}$ and $W_{f}$ are the height and width of the filter parameter matrix $F$ between $L_{l-1}$ and $L_{l}$.
Based on the error set of neurons, the weight parameters of $F$ are computed subsequently.
The gradient of each weight $w_{i,j}$ is calculated in parallel, as defined in Eq. (\ref{eq23}):
{\small
\begin{equation}
\setlength{\abovecaptionskip}{0pt}
\setlength{\belowcaptionskip}{0pt}
\label{eq23}
\frac{\partial E_{d}}{\partial w_{i,j}} =\sum_{h=1}^{H_{f}}\sum_{m=1}^{W_{f}}{\delta_{h,m}^{l} \times a_{i+h,j+m}^{l-1}}.
\end{equation}
The} gradient of the bias weight $w_{b}$ is computed in Eq. (\ref{eq24}):
{\small
\begin{equation}
\setlength{\abovecaptionskip}{0pt}
\setlength{\belowcaptionskip}{0pt}
\label{eq24}
\frac{\partial E_{d}}{\partial w_{b}} = \sum_{a_{i} \in L_{l-1}}\sum_{a_{j} \in L_{l}}{\delta_{i,j}^{l}}.
\end{equation}
Based} on the gradient values, each weight $w_{i,j}$ is updated in Eq. (\ref{eq25}):
{\small
\begin{equation}
\setlength{\abovecaptionskip}{0pt}
\setlength{\belowcaptionskip}{0pt}
\label{eq25}
w_{i,j} = w_{i,j} - \eta \frac{\partial E_{d}}{\partial w_{i,j}},
\end{equation}
where} $\eta$ is the learning rate of the CNN network.

\subsection{Implementation of Inner-layer Parallel Training}
\label{section4.3}

\begin{figure}[!htb]
 \setlength{\abovecaptionskip}{0pt}
 \setlength{\belowcaptionskip}{0pt}
 \centering
 \includegraphics[width=3.4in]{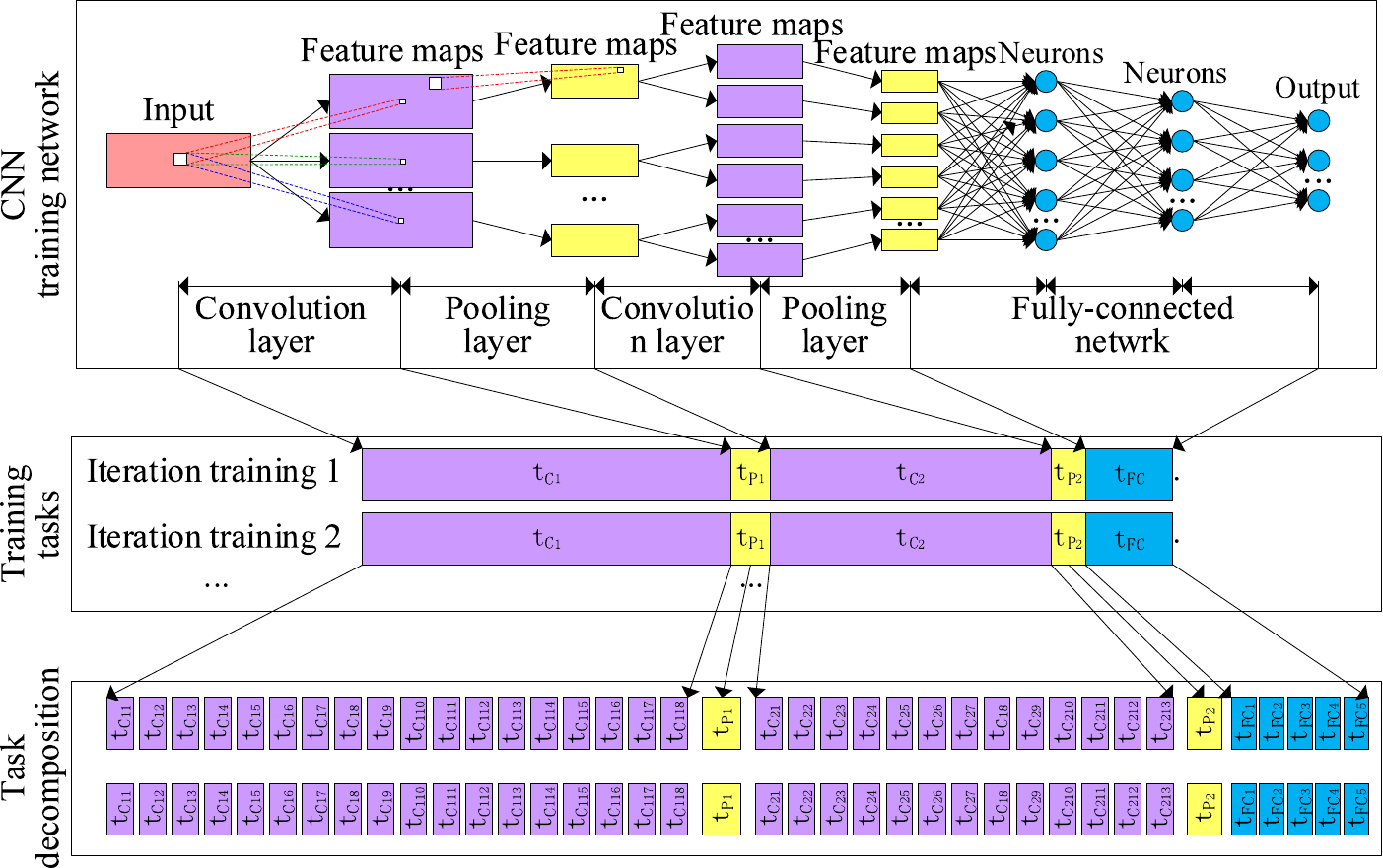}
 \caption{Task decomposition for a CNN subnetwork.}
 \label{fig09}
\end{figure}

We implement the inner-layer parallel training of BPT-CNN on computing nodes equipped with multi-core CPUs.
Based on the parallel models proposed in the previous section, computing tasks of these training phases are decomposed into several subtasks.
The workflow of task decomposition for a CNN subnetwork is illustrated in Fig. \ref{fig09}.

(1) Task priority marking.

According to the logical and data dependence of the decomposed subtasks, a task Directed Acyclic Graph (DAG) is created.
With the thread-level load balancing and completion time minimization as the optimization goal, the priorities of tasks in the task DAG are marked.
We set a maximum value for the entrance task of the task DAG graph.
Then, the priorities of tasks in each level are set according to the tasks' level.
Specifically, upstream tasks' priorities are higher than that of downstream tasks, while tasks at the same level have the same priority.

(2) Task scheduling and execution.

\begin{figure}[ht]
 \setlength{\abovecaptionskip}{0pt}
 \setlength{\belowcaptionskip}{0pt}
 \centering
 \includegraphics[width=3.4in]{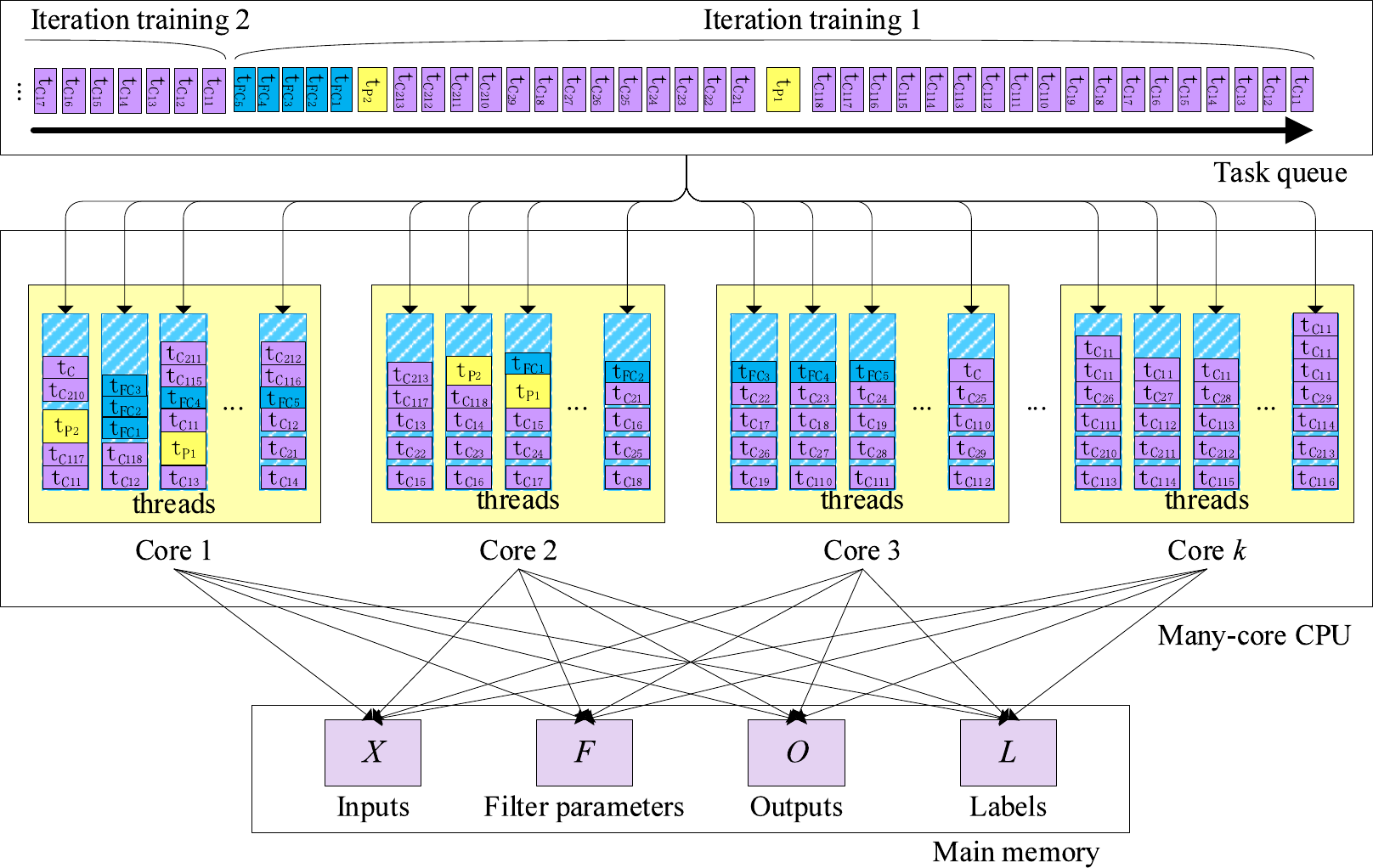}
 \caption{Task scheduling of a CNN subnetwork on multi-core CPU platform.}
 \label{fig10}
\end{figure}

Based on the priorities of tasks, we allocate these tasks to threads on the multi-core CPU platform using the priority task scheduling algorithm \cite{d43}.
Based on the task priorities, tasks of the entire CNN training network are allocated to different threads on the different CPU cores.
An example of the task scheduling of the CNN training network with multi-threaded parallelism is illustrated in Fig. \ref{fig10}.

\begin{algorithm}[!ht]
\scriptsize
\caption{{\scriptsize Parallel task scheduling of BPT-CNN.}}
\label{alg07}
\begin{algorithmic}[1]
\REQUIRE ~\\
$PTs$: The list of parallel tasks $PTs$ for the current CNN network;\\
$Ths$: the available threads on the current computing nodes.\\
\ENSURE ~\\
$A(PTs)$: a schedule of tasks in $PTs$ which maximizes thread-level load balancing and minimizes the waiting time of critical paths.\\
\STATE order $PTs$ with priority level $\leftarrow$ orderByPriority($PTs$);
\STATE \textbf{while} $PTs \neq NULL$ \textbf{do}
\STATE \quad take the top element from the task list $t_{i}$ $\leftarrow$ $PTs$.top();
\STATE \quad get the logical or data depended tasks $ts_{i}$ of $t_{i}$;
\STATE \quad \textbf{for} $t_{j}$ in $ts_{i}$ \textbf{do}
\STATE \qquad \textbf{if} $t_{j}$.state $\neq$ complete \textbf{do}
\STATE \qquad \quad $t_{i}$ waits and break;
\STATE \quad find thread $Th_{k}$ from $THs$ with minimal workload;
\STATE \quad call Assignment $A(PTs)$ $\leftarrow$ ($t_{i}$, $Th_{k}$);
\STATE \quad remove $t_{i}$ from $PTs$;
\RETURN $A(PTs)$.
\end{algorithmic}
\end{algorithm}

\section{Experiments}
\label{section6}

\subsection{Experimental Settings}
\label{section5.1}
All of the experiments are conducted on a distributed computing cluster built with 30 high-performance computing nodes, and each of them is equipped with Intel Xeon Nehalem EX CPU and 48 GB main memory, respectively.
Each Nehalem-EX processor features up to 8 cores inside a single chip supporting 16 threads and 24MB of cache.
Comparison experiments are conducted to evaluate the proposed BPT-CNN by comparing with Tensorflow CNN \cite{d15}, DisBelief \cite{d02}, and DC-CNN \cite{d10} algorithms, in terms of accuracy and performance evaluation.
Large-scale public image datasets from ImageNet \cite{d22} with 14,197,122 samples are used in the experiments.

\subsection{Accuracy Evaluation}
\label{section5.2}

\begin{figure}[!ht]
\setlength{\abovecaptionskip}{0pt}
\setlength{\belowcaptionskip}{0pt}
\centering
 \subfigure[Accuracy]{\includegraphics[width=1.6in]{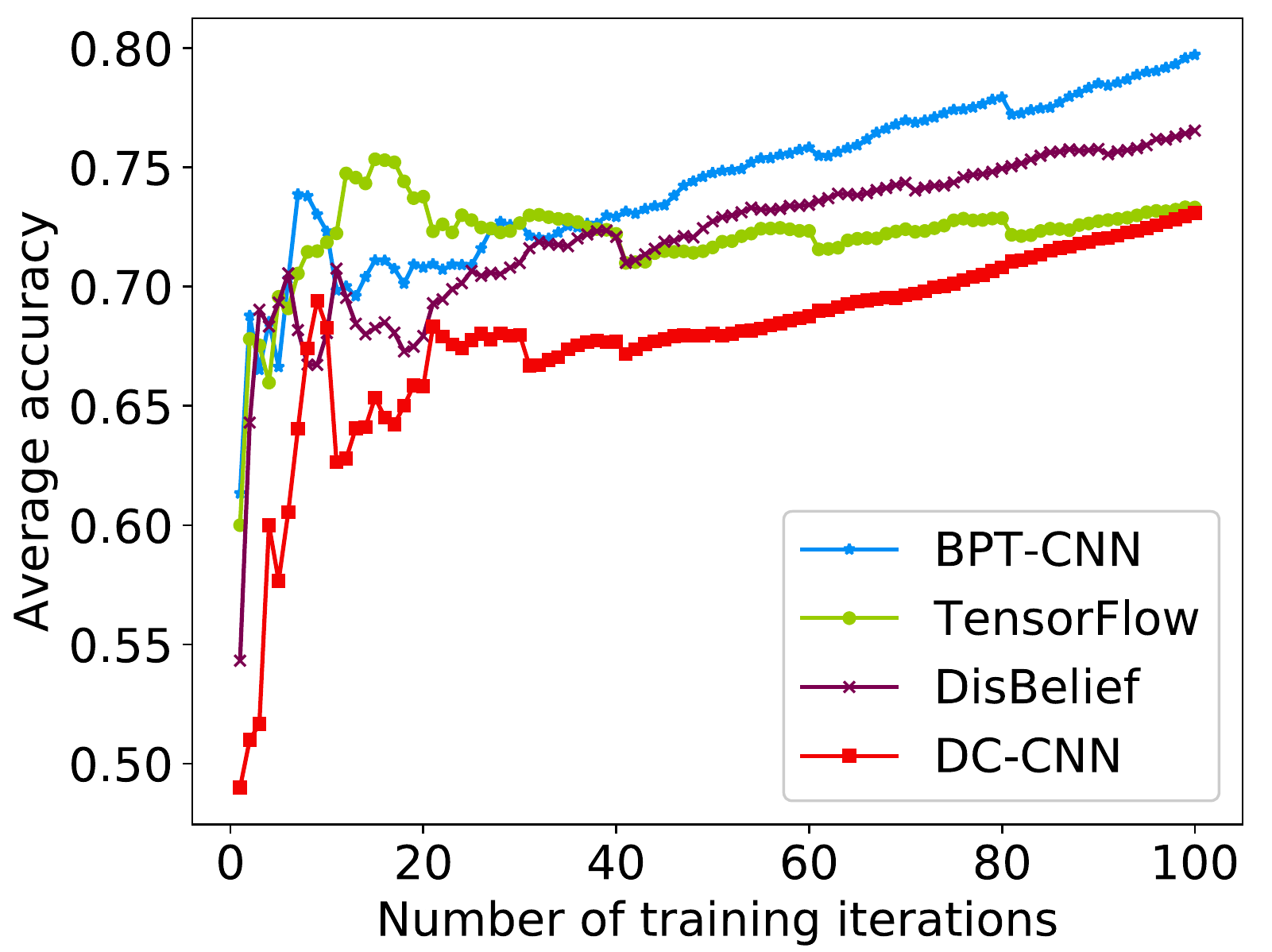}}
 \subfigure[AUC]{\includegraphics[width=1.6in]{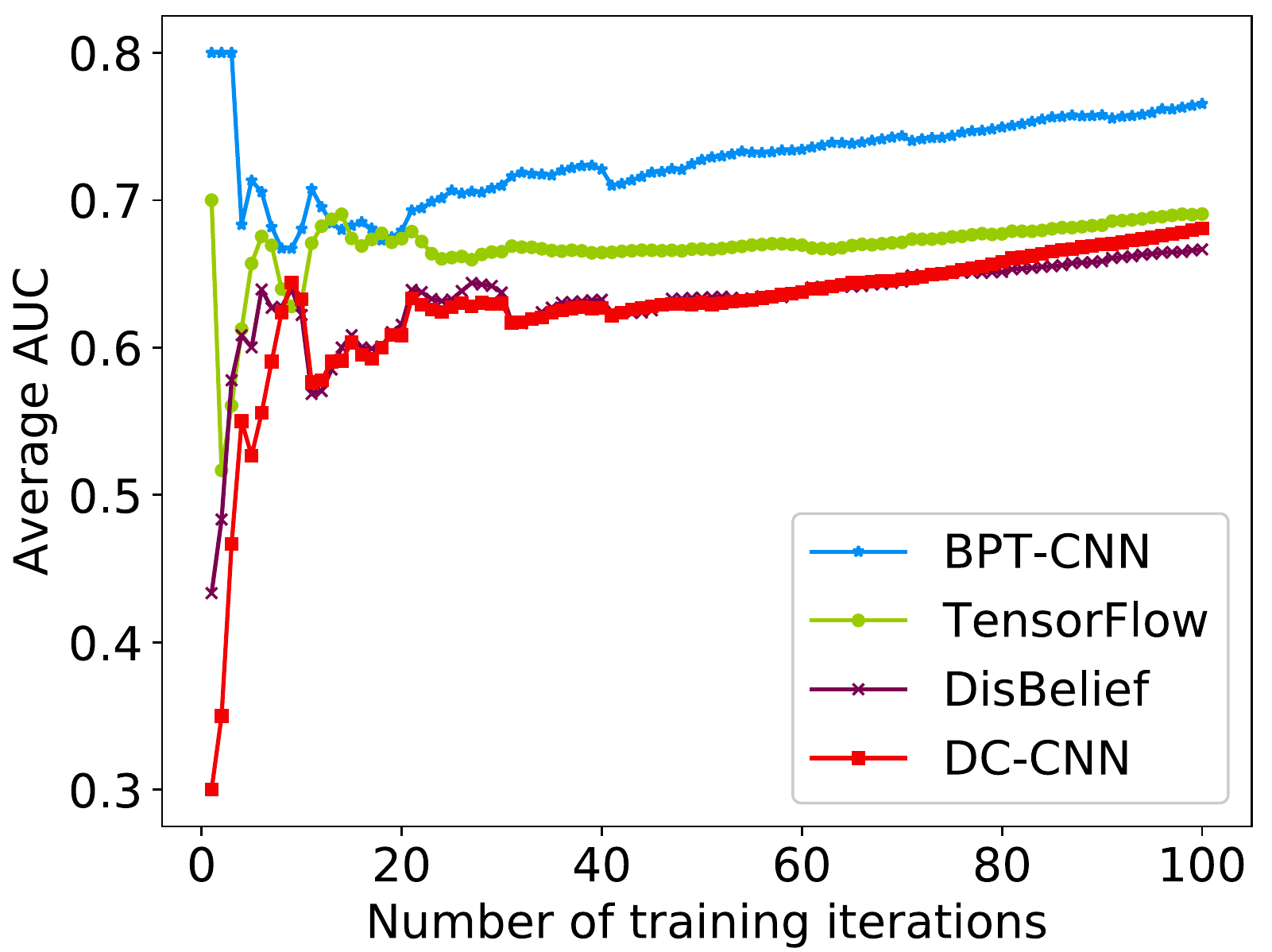}}
\caption{Accuracy evaluation of the comparison algorithms.}
\label{chart01}
\end{figure}

We evaluate the accuracy of BPT-CNN by comparing with Tensorflow, DisBeilef, and DC-CNN.
For each algorithm, five-fold experiments on the ImageNet dataset with 100 epoch iterations are conducted and the average values of accuracy and the Area Under the Curve (AUC) are compared.
The experimental results of accuracy and AUC of the comparison algorithms are presented in Fig. \ref{chart01}.

As shown in Fig. \ref{chart01} (a) and (b), BPT-CNN achieves the similar accuracy with compared algorithms, as well as higher AUC values in most of the cases.
The average value of accuracy of BPT-CNN is equal to 0.744, while that of Tensorflow, DisBelief, and DC-CNN is 0.721, 0.722, and 0.639, respectively.
Because of the parallel training and global weight updating, BPT-CNN narrows the impact of local overfitting and obtains more stable and robust global network weights.
As the epoch of iteration training increases, both of accuracy and AUC of BPT-CNN steadily increases.
AUC of BPT-CNN is greater than that of Tensorflow by 5.91\%, on average, 9.56\% higher than that of DisBelief, and 10.09\% higher than that of DC-CNN.
Therefore, compared with Tensorflow, DisBelief, and DC-CNN, BPT-CNN does not reduce the accuracy of CNNs.
Moreover, benefitting from the global weight updating strategy, BPT-CNN achieves more robustness than compared algorithms.

\subsection{Performance Evaluation}
\label{section5.3}

\subsubsection{Execution Time of Comparison Algorithms}
\begin{figure}[!ht]
\setlength{\abovecaptionskip}{0pt}
\setlength{\belowcaptionskip}{0pt}
\centering
 \subfigure[Different data sizes]{\includegraphics[width=1.6in]{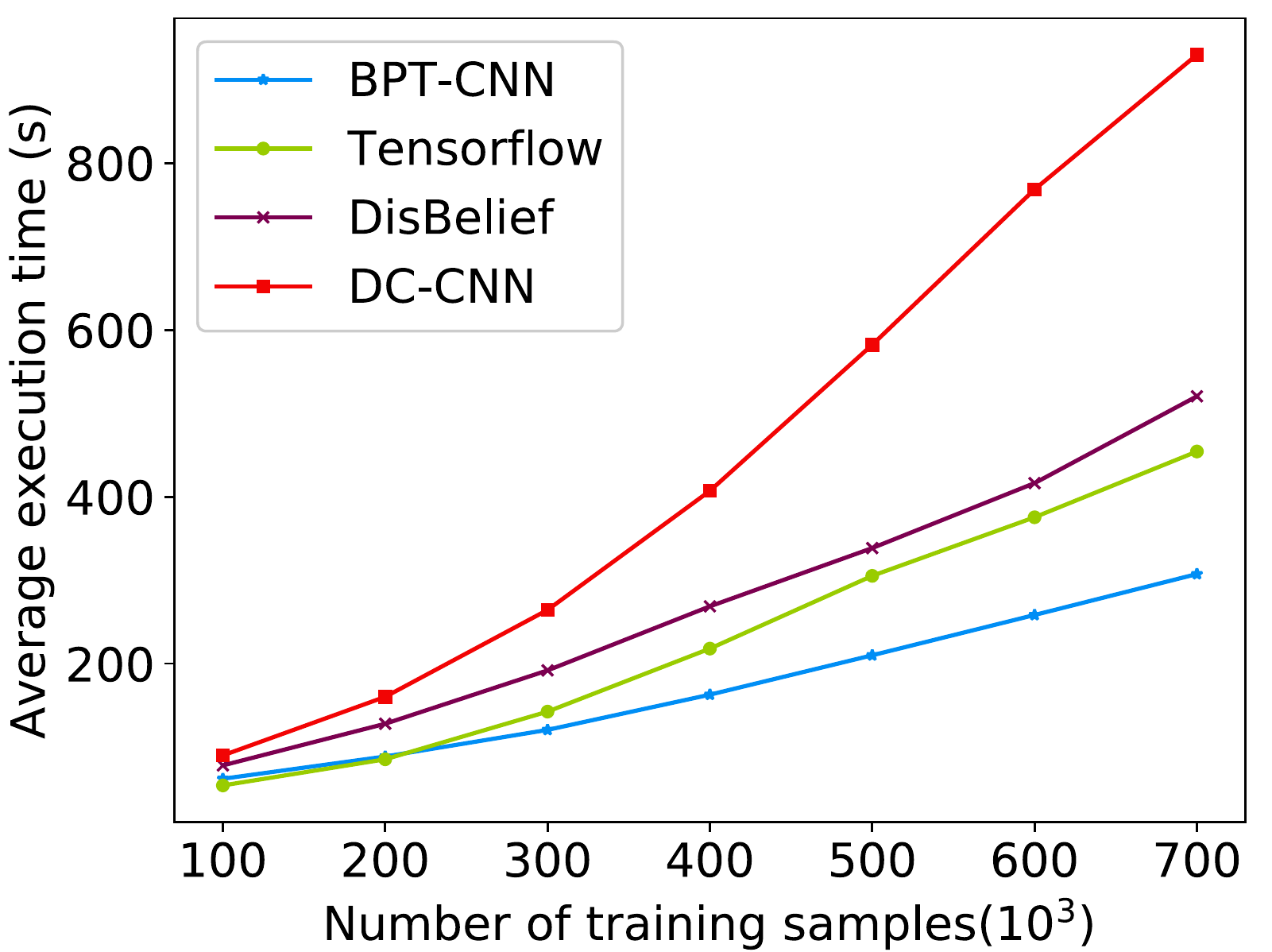}}
 \subfigure[Different cluster scales]{\includegraphics[width=1.6in]{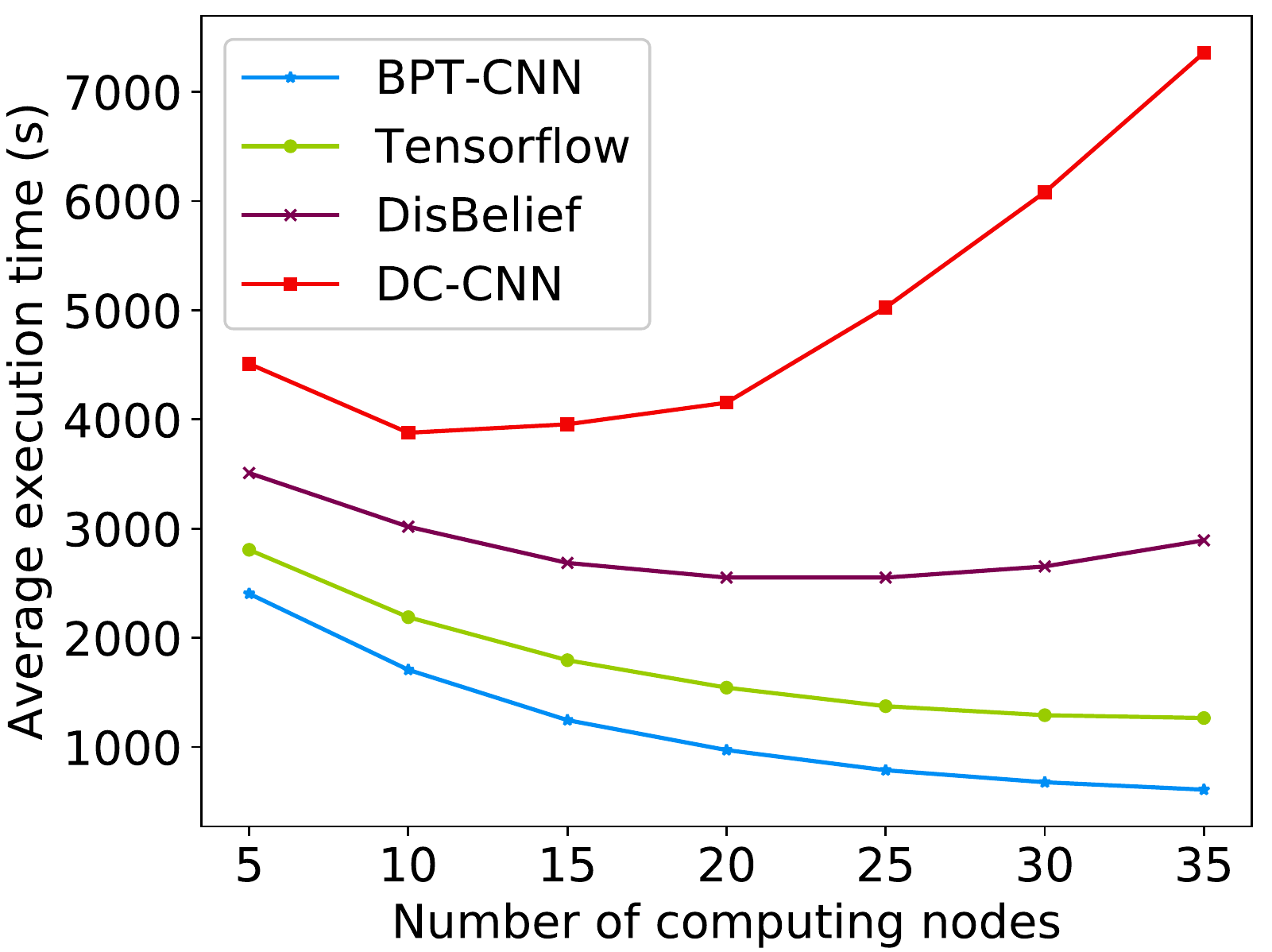}}
\caption{Total execution time of the comparison algorithms.}
\label{chart02}
\end{figure}

The execution time of these algorithms is compared using 100 training iterations in various configurations: different data sizes and computing cluster scales.
The comparison of the average execution time of each algorithm in each case is shown in Fig. \ref{chart02}.

As can be seen in Fig. \ref{chart02} (a) and (b), the proposed BPT-CNN algorithm achieves higher performance than the compared algorithms in most of the cases.
Benefitting from the data-parallelism strategy, when the data size increases, the volume of each partitioned subset on each computer is slightly increased, leading to a slight increase in the average workload of each computer.
For example, when the number of training samples increases from 100,000 to 700,000, the execution time of BPT-CNN rises from 62.77s to 307.35s, while that of Tensorflow increases from 54.38s to 454.23s, and that of DC-CNN sharply increases from 91.21s to 929.74s.
In addition, taking advantage of the IDPA strategy, the proposed BPT-CNN algorithm owns scalability over the compared algorithms.
When the scale of the computing cluster expended, the execution time of BPT-CNN and Tensorflow is significantly reduced.
Experimental results indicate that BPT-CNN achieves high performance and scalability in distributed computing clusters.

\subsubsection{Execution Time Comparison for Fixed Accuracy}
Considering the different training architectures of various comparison algorithms, we discuss how these algorithm trade off  performance and accuracy with resource consumption.
We discuss the training iterations required for each algorithm to achieve different accuracy, and then measure the execution time each algorithm takes under different computing resources.
The comparison results are shown in Table \ref{table53} and Fig. \ref{chart03}.

\begin{table}[!ht]
\scriptsize
\centering
\renewcommand{\arraystretch}{1.0}
\setlength{\abovecaptionskip}{0pt}
\setlength{\belowcaptionskip}{0pt}
\caption{{\scriptsize Training iterations required by comparison algorithm for different accuracy.}}
\label{table53}
\tabcolsep1pt
\begin{tabular}{C{1.0cm} C{1.7cm} C{1.7cm} C{1.7cm} C{1.7cm}}
\hline
\textbf{Accuracy} & \textbf{BPT-CNN} & \textbf{Tensorflow} & \textbf{DisBelief} & \textbf{DC-CNN}\\
\hline
0.650	 & 7 & 7 & 9 & 12 \\
0.700     & 18 & 15 & 22 & 28\\
0.750   & 42 & 64 & 85 &147\\
0.800      & 97 & 187 & 211 & -\\
\hline
\end{tabular}
\end{table}

\begin{figure}[!ht]
\setlength{\abovecaptionskip}{0pt}
\setlength{\belowcaptionskip}{0pt}
\centering
 \subfigure[Different data sizes]{\includegraphics[width=1.6in]{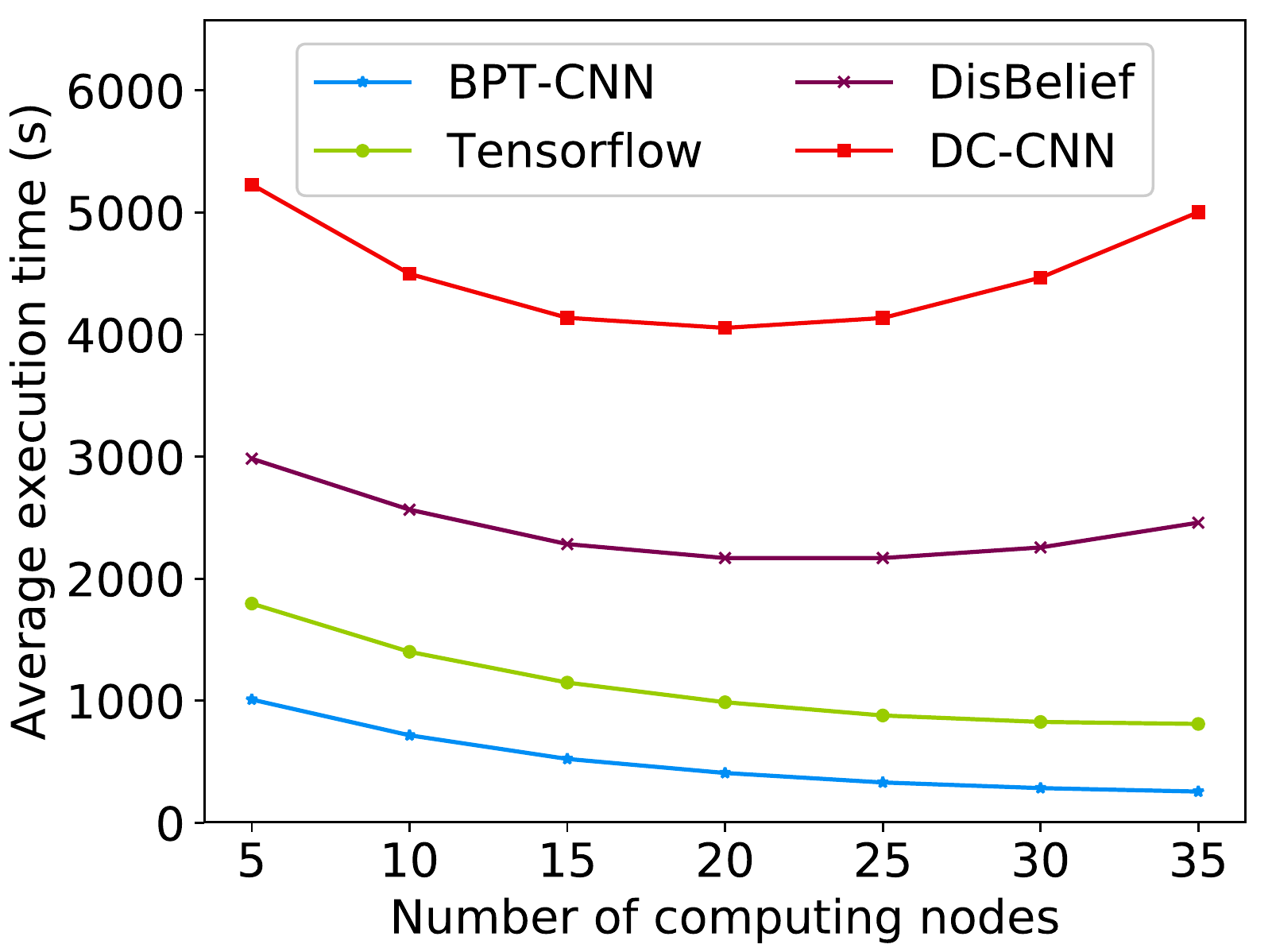}}
 \subfigure[Different cluster scales]{\includegraphics[width=1.6in]{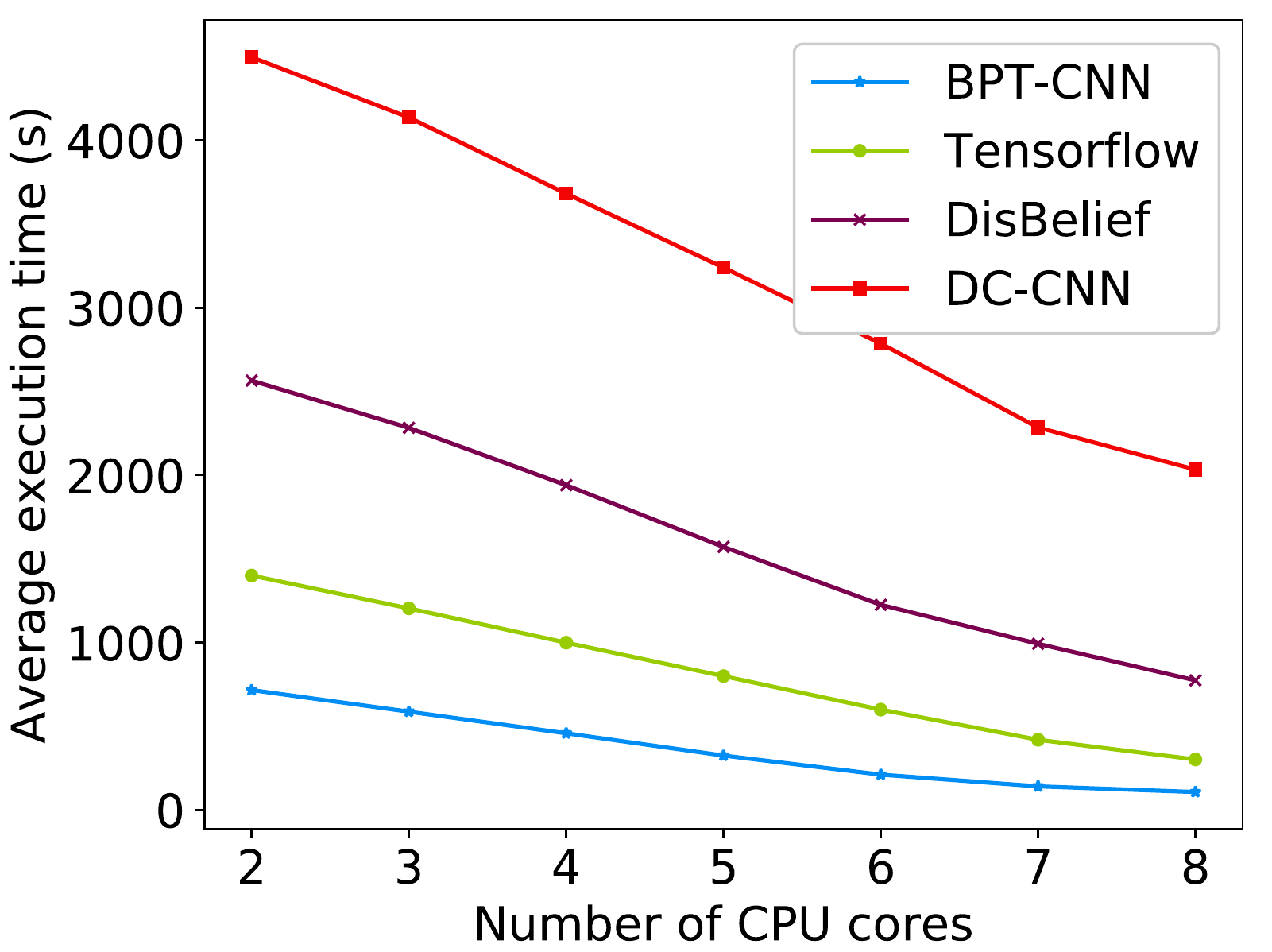}}
\caption{Total execution time of the comparison algorithms.}
\label{chart03}
\end{figure}

From Table \ref{table53}, all algorithms use similar iterations to achieve an accuracy of 0.650.
However, to achieve higher accuracy, BPT-CNN requires fewer iterations than Tensorflow, DisBelief, and DC-CNN.
For example, BPT-CNN requires 42 iterations to achieve an accuracy of 0.750, while Tensorflow uses 64 and DisBelief uses 85, and DC-CNN requires up to 147.
In addition, to achieve an accuracy of 0.750, we compare the actual execution times of each algorithm under different numbers of computing nodes and CPU cores, as shown in Fig. \ref{chart03} (a) and (b).
When the scale of the computing cluster and CPU cores expended, the execution time of BPT-CNN and Tensorflow is significantly reduced.
In contrast, the execution time of DisBelief and DC-CNN algorithms is increased when the cluster scale reaches a certain amount (e.g., 25-35), which is caused by the more data communication among the increasing machines.
Experimental results indicate that BPT-CNN achieves higher accuracy and performance than other algorithms using the same computing resource.
Moreover, when the scale of computing nodes and CPU cores increases, the performance benefits of BPT-CNN is more noticeable.

\subsubsection{{\small Execution Time of BPT-CNN with Different Strategies}}
We further evaluate the performance of the proposed BPT-CNN under different global weight update and data partitioning strategies.
To evaluate the effectiveness of the IDPA strategy, we perform the same work using the Uniform Data Partitioning and Allocation (UDPA) strategy, where the training dataset is uniformly partitioned into $m$ partitions and allocated into the $m$ computers.
Comparison experiments are conducted in terms of data size, computing cluster scale, CNN network scale, and thread size.
The average execution time of BPT-CNN with different strategies is presented in Fig. \ref{chart04}.

In Fig \ref{chart04} (a), 7 different scales of CNN network are constructed in the experiments, as described in Table \ref{table52}.
Here ``layers(Conv)'' and ``filters(Conv)'' denote the number of the convolutional layer and that of filters at each layer, respectively.
``layers(FC)'' and ``neurons(FC)'' denote the number of layers in the fully-connected layers and number of neurons in each layer, respectively.

\begin{table}[!ht]
\scriptsize
\centering
\renewcommand{\arraystretch}{1.0}
\setlength{\abovecaptionskip}{0pt}
\setlength{\belowcaptionskip}{0pt}
\caption{{\scriptsize Different scales of CNN network used in the experiments.}}
\label{table52}
\tabcolsep1pt
\begin{tabular}{L{2.0cm} C{0.85cm} C{0.85cm} C{0.85cm} C{0.85cm} C{0.85cm} C{0.85cm} C{0.85cm}}
\hline
\textbf{Scales} & \textbf{case1} & \textbf{case2} & \textbf{case3} & \textbf{case4} & \textbf{case5} & \textbf{case6} & \textbf{case7}\\
\hline
layers(Conv) & 2 & 4 & 6 & 8 & 8 & 10 & 10 \\
filters(Conv)& 4 & 4 & 8 & 8 & 10 & 10 & 12\\
layers(FC)   & 3 & 3 & 5 & 5 & 7 & 7 & 7\\
neurons(FC)  & 500 & 1000 & 1500 & 1500 & 2000 &  2000 & 2000\\
\hline
\end{tabular}
\end{table}

\begin{figure}[!ht]
\setlength{\abovecaptionskip}{0pt}
\setlength{\belowcaptionskip}{0pt}
\centering
 \subfigure[Different data sizes]{\includegraphics[width=1.6in]{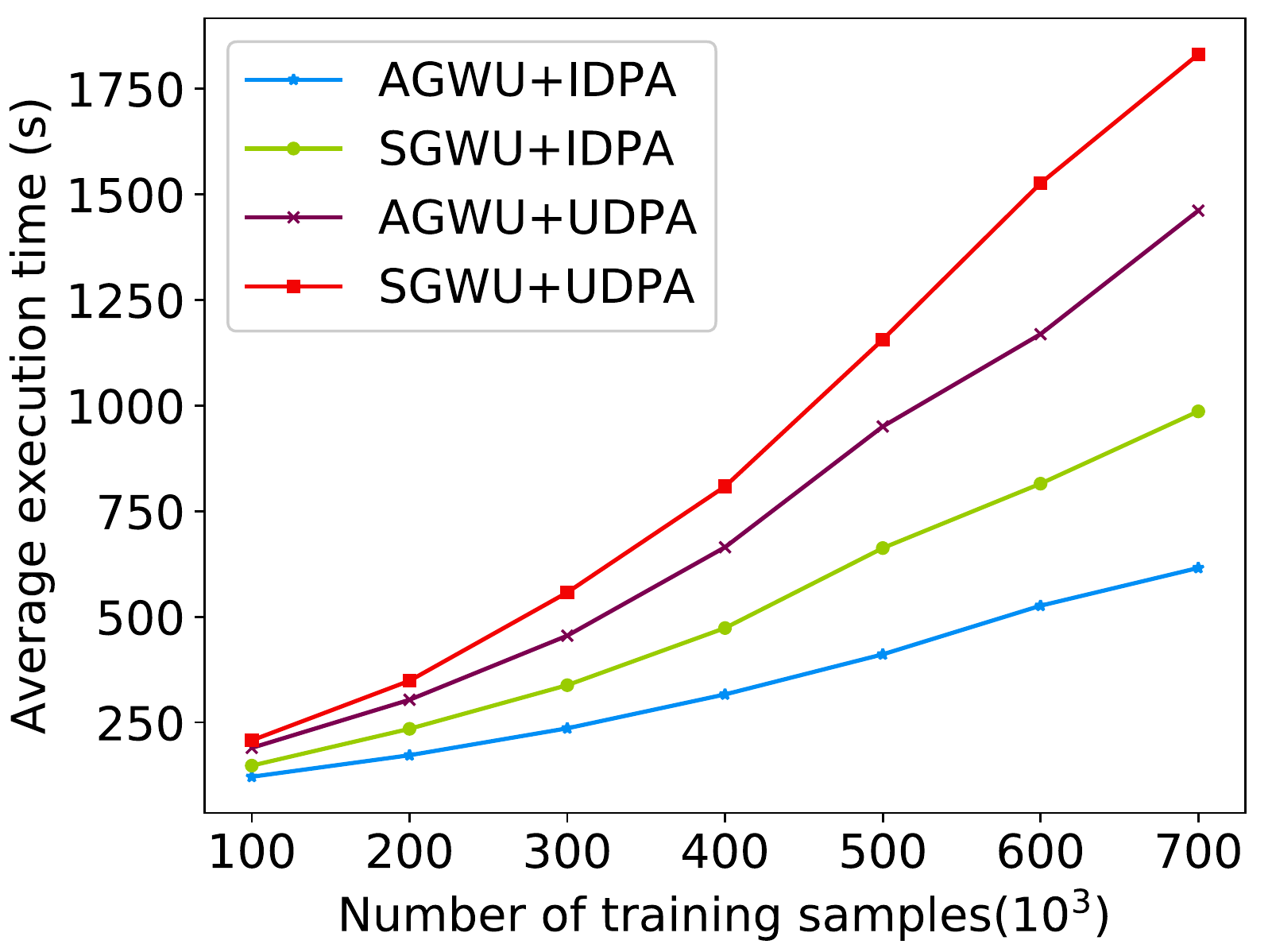}}
 \subfigure[Different network scales]{ \includegraphics[width=1.6in]{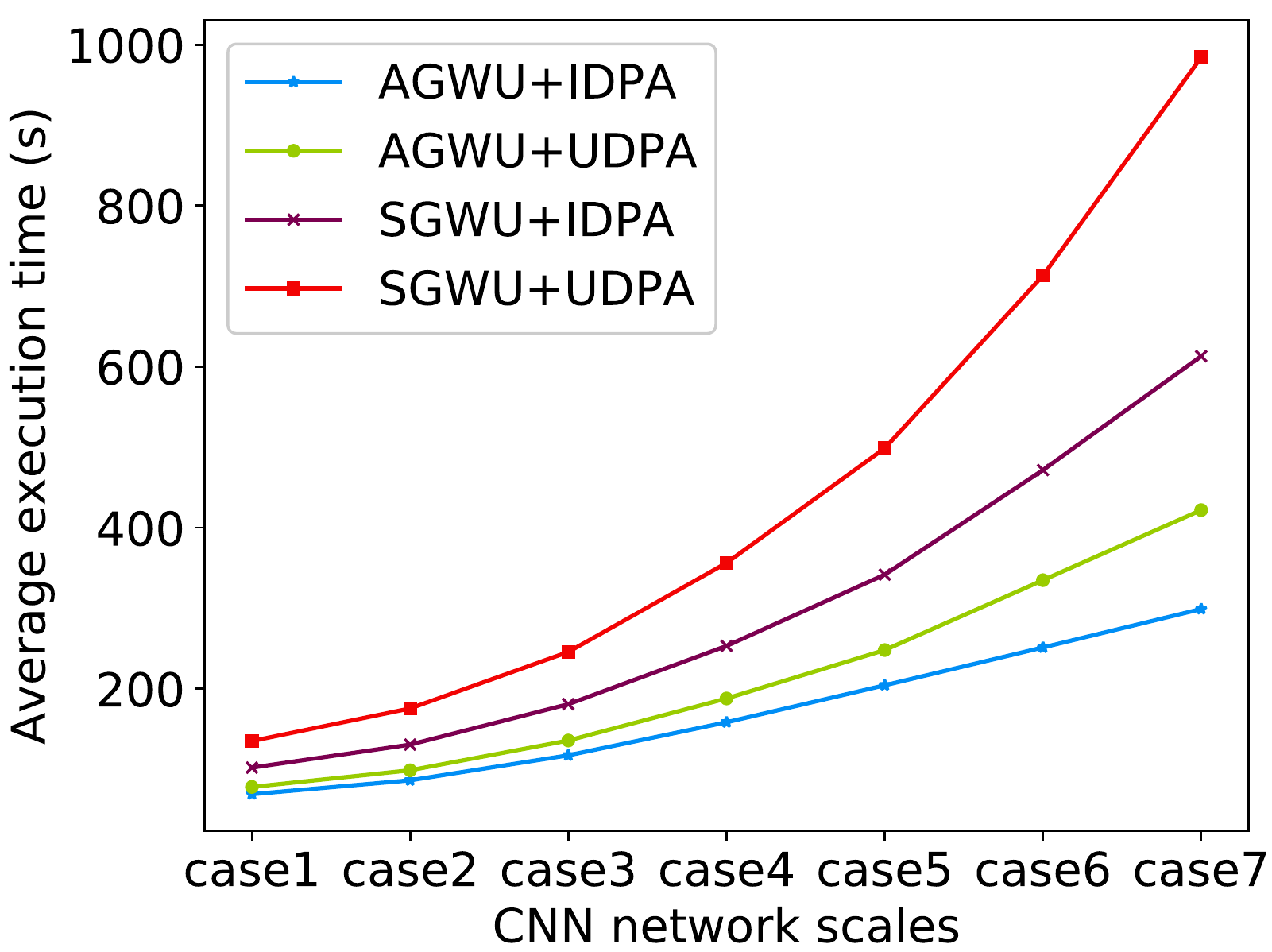}}
  \subfigure[Different computing nodes]{ \includegraphics[width=1.6in]{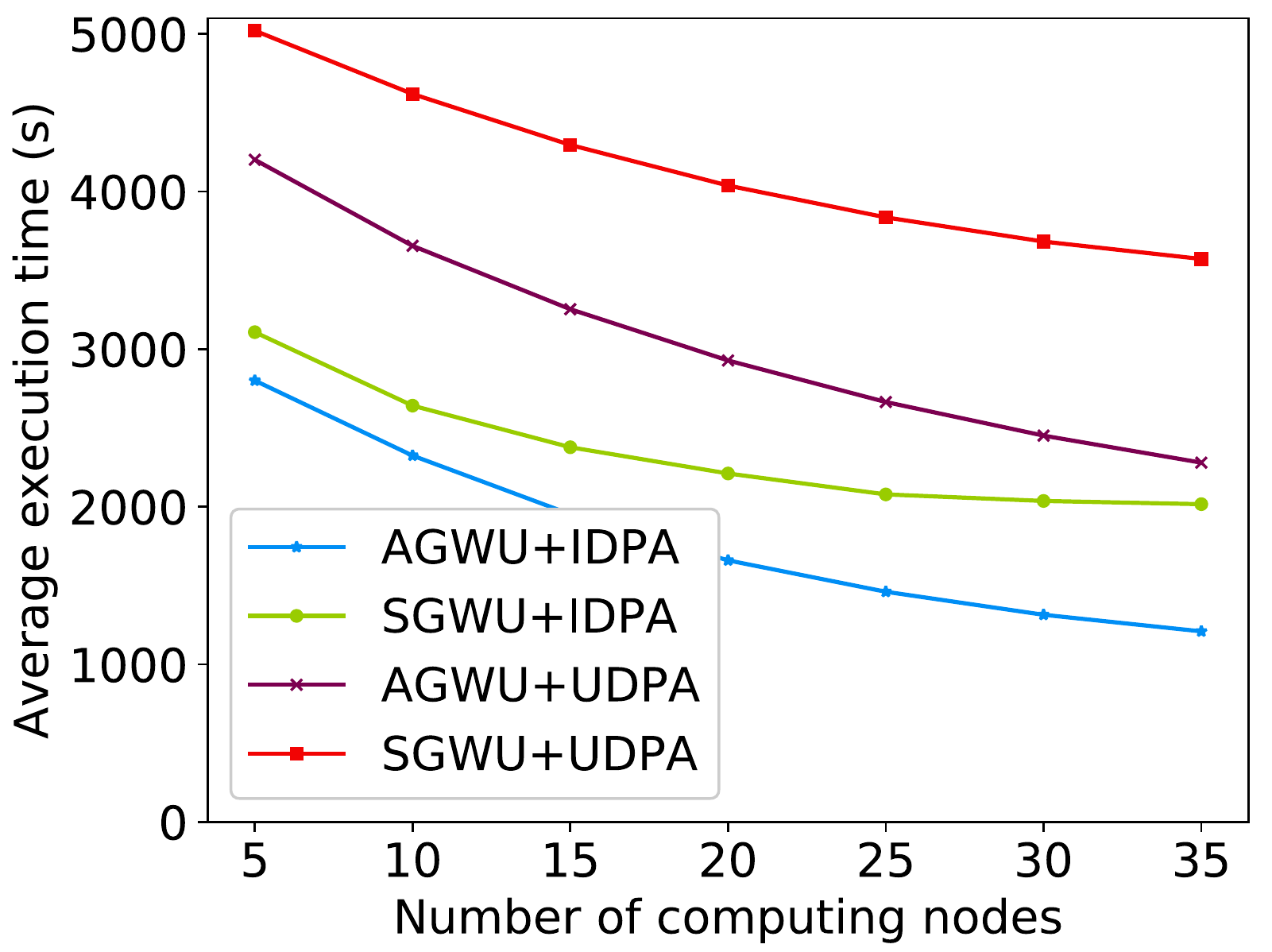}}
  \subfigure[Different thread scales]{ \includegraphics[width=1.6in]{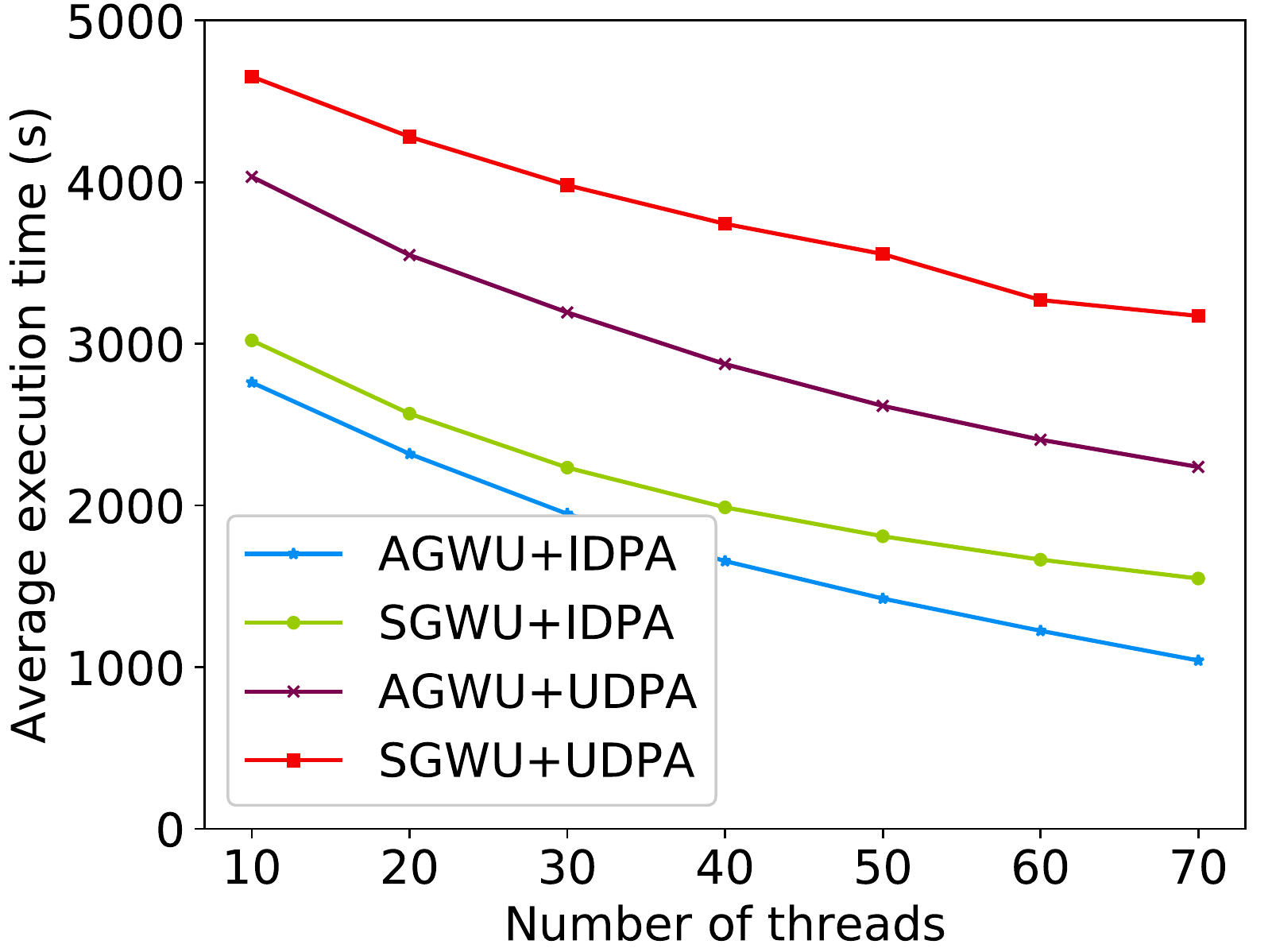}}
\caption{Execution time of BPT-CNN with different strategies.}
\label{chart04}
\end{figure}

By comparing strategies AGWU and SGWU, the execution time of BPT-CNN using AGWU is obviously lower than SGWU in most cases.
In AGWU, because of the asynchronous update of the global weight set, each computing node uses the minimum time to wait for the global weight update and trains almost continuously.
In addition, by comparing data partitioning strategies IDPA and UDPA, benefitting from the incrementa data partitioning, the workload of computing nodes stays well balanced, which further shortens the waiting time among different machines.
As shown in each case in Fig. \ref{chart04}, the execution time of BPT-CNN with IDPA is significantly lower than that with UDPA strategy.
Hence, BPT-CNN using AGWU+IDPA strategies exhibits the most efficient performance against other cases.
Moreover, with the increase of data size or CNN network scale, the execution time of BPT-CNN using AGWU+IDPA maintains a slow rise.
When the computing cluster scale and the number of threads on each machine increases, the benefits of AGWU+IDPA are more noticeable.
Taking advantage of the IDPA and AGWU strategies, BPT-CNN achieves significant strength in terms of performance.

\subsection{Data Communication and Workload Balancing}
\label{section5.4}
\begin{figure}[!ht]
\setlength{\abovecaptionskip}{0pt}
\setlength{\belowcaptionskip}{0pt}
\centering
 \subfigure[Impact of data size]{\includegraphics[width=1.6in]{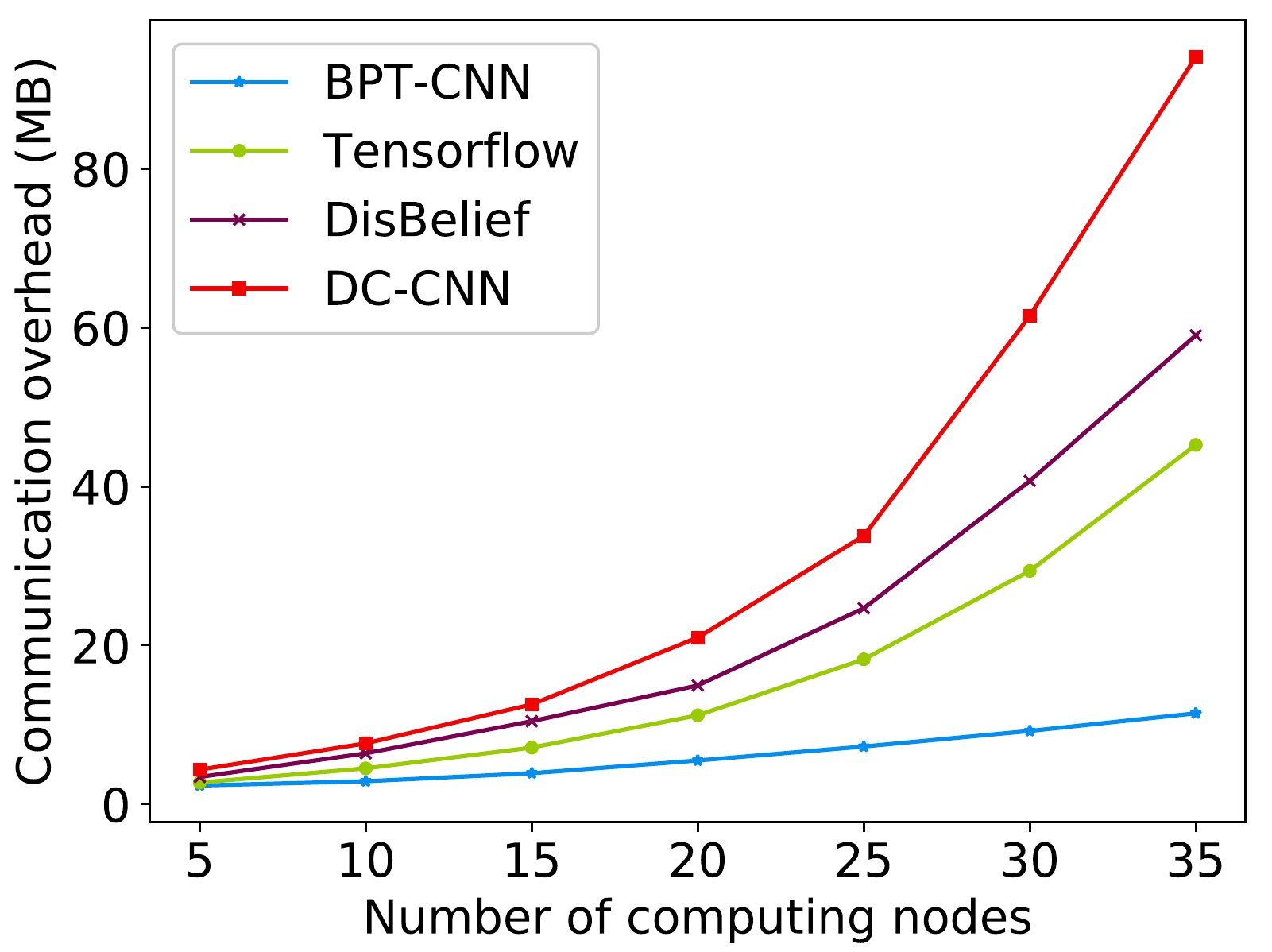}}
 \subfigure[Impact of cluster scale]{\includegraphics[width=1.6in]{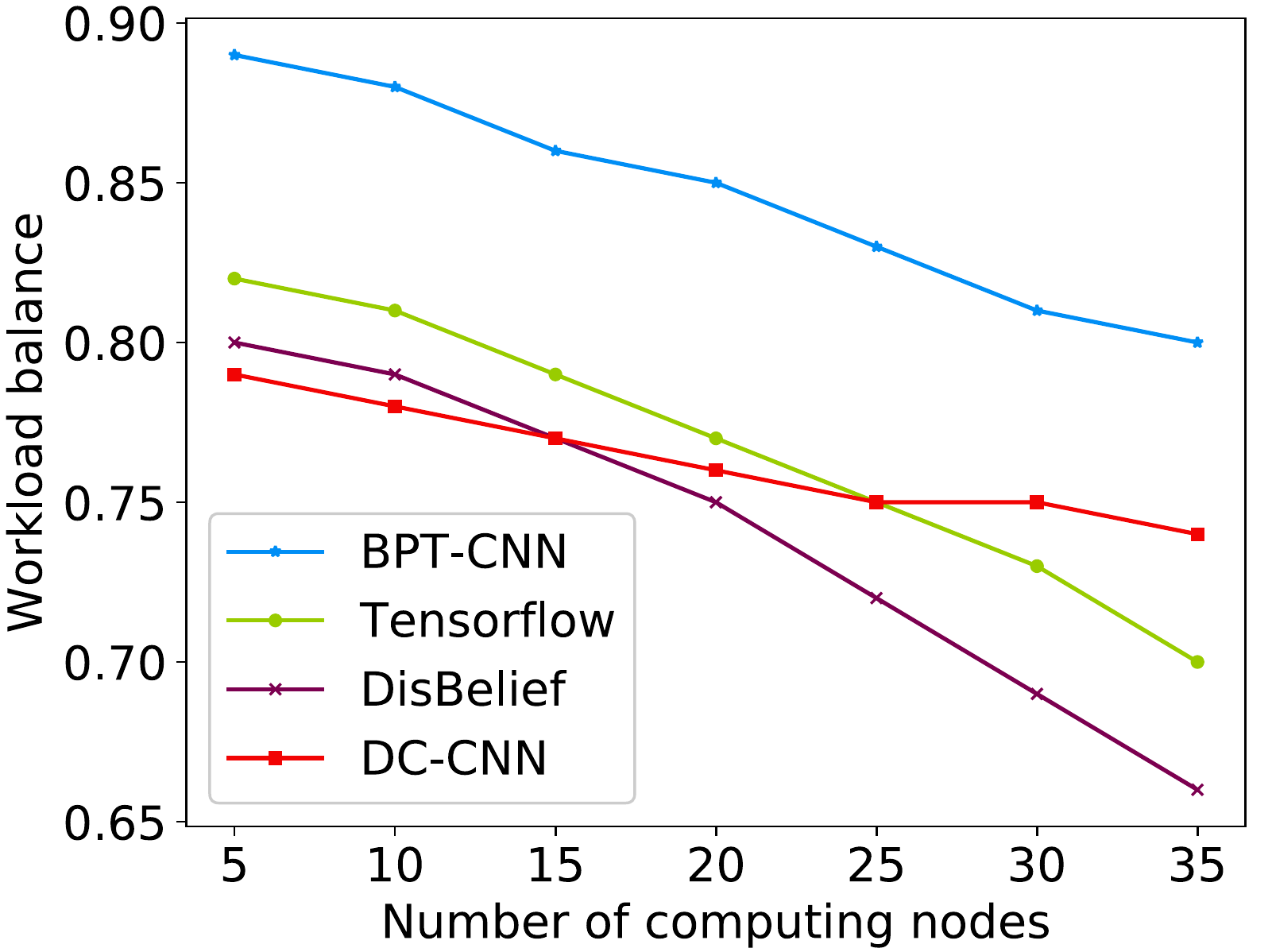}}
\caption{Comparison on data communication and workload balancing.}
\label{chart05}
\end{figure}

We evaluate the proposed BPT-CNN architecture in the view of data communication overhead and workload balancing by comparing with Tensorflow, DisBeilef, and DC-CNN algorithms.
600,000 training samples are used in the experiments, and the number of computing nodes increases from 5 to 35 in each case.
Experiment results of data communication and workload balancing are shown in Fig. \ref{chart05}.

It is clear from Fig. \ref{chart05} (a) and (b) that, in most cases, BPT-CNN owns significant workload balancing and lower data communication costs than other algorithms.
Due to the use of the IDPA strategy in BPT-CNN, there is only communication overhead between the computing nodes for transmitting local/global weight parameters, and no training sample migration is required.
Hence, as the number of computing nodes increases from 5 to 35, the communication overhead of BPT-CNN slowly increases from 2.35 MB to 11.44 MB.
In contrast, due to dynamic resource scheduling, Tensorflow generates communication overhead from 2.73 MB for 5 computers to 45.23 MB between 35 computers.
Moreover, to achieve workload balancing, DisBelief and DC-CNN use data migration operations during training, which results in heavy communication overhead between computers.

We compare the workload balance of each algorithm under different scales of the computing cluster, as shown in Fig. \ref{chart05} (b).
Our BPT-CNN architecture considers the heterogeneity of compute nodes and allocates corresponding workloads based on the actual computing power of each compute node.
Hence, as the scale of the cluster increases, BPT-CNN achieves a stable workload balance, keeping between 0.89 and 0.80.
In contrast, without heterogeneity-aware data allocation, the workload of other comparison algorithms is not as balanced as BPT-CNN.
The unbalanced workload further leads to long waiting time for synchronization and more execution time for the entire CNN network.
Experimental results demonstrate that BPT-CNN significantly improves the workload balance of the distributed computing cluster with acceptable communication overhead.

\section{Conclusions}
\label{section7}
This paper presented a bi-layered parallel training architecture to accelerate the training process of large-scale CNNs.
In the outer-layer parallel training, the performance of the entire CNN network is significantly improved based on data-parallelism optimization, where the issues of data communication, workload balance, and synchronization, are well addressed.
In the inner-layer parallelism, the training process of each CNN subnetwork is further accelerated using task-parallelism optimization.
Extensive experimental results on large-scale datasets indicate that the proposed BPT-CNN effectively improves the training performance of CNNs in distributed computing clusters with minimum data communication and synchronization waiting.

For future work, we will further concentrate on scalable CNN models and the parallelization of deep learning algorithms on high-performance computers.
In addition, development of deep learning algorithms specific applications is also an interesting topic, such as scalable CNNs for images and LSTMs for time series.

\section*{Acknowledgment}
This research is partially funded by the National Key R\&D Program of China (Grant No. 2016YFB0200201),
the Key Program of the National Natural Science Foundation of China (Grant No. 61432005),
the National Outstanding Youth Science Program of National Natural Science Foundation of China (Grant No. 61625202),
the International Postdoctoral Exchange Fellowship Program (Grant No. 2018024),
and the China Postdoctoral Science Foundation funded project (Grant No. 2018T110829).
This work is also supported in part by NSF through grants IIS-1526499, IIS-1763325, CNS-1626432, and NSFC 61672313.

\ifCLASSOPTIONcaptionsoff
\newpage
\fi

\bibliographystyle{IEEEtran}
\bibliography{reference}

\begin{IEEEbiography}
[{\includegraphics[width=1in, height=1.25in, clip, keepaspectratio]{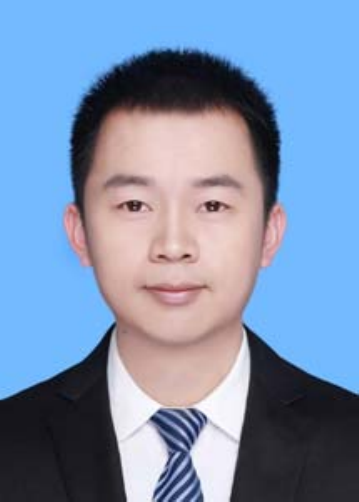}}]
{Jianguo Chen}  received the Ph.D. degree in College of Computer Science and Electronic Engineering at Hunan University, China.
He was a visiting Ph.D. student at the University of Illinois at Chicago from 2017 to 2018.
He is currently a postdoctoral in University of Toronto and Hunan University.
His major research areas include parallel computing, cloud computing, machine learning, data mining, bioinformatics and big data.
He has published research articles in international conference and journals of data-mining algorithms and parallel computing, such as
{\em IEEE TPDS}, IEEE/ACM TCBB, and {\em Information Sciences}.
\end{IEEEbiography}

\begin{IEEEbiography}
[{\includegraphics[width=1in, height=1.25in, clip, keepaspectratio]{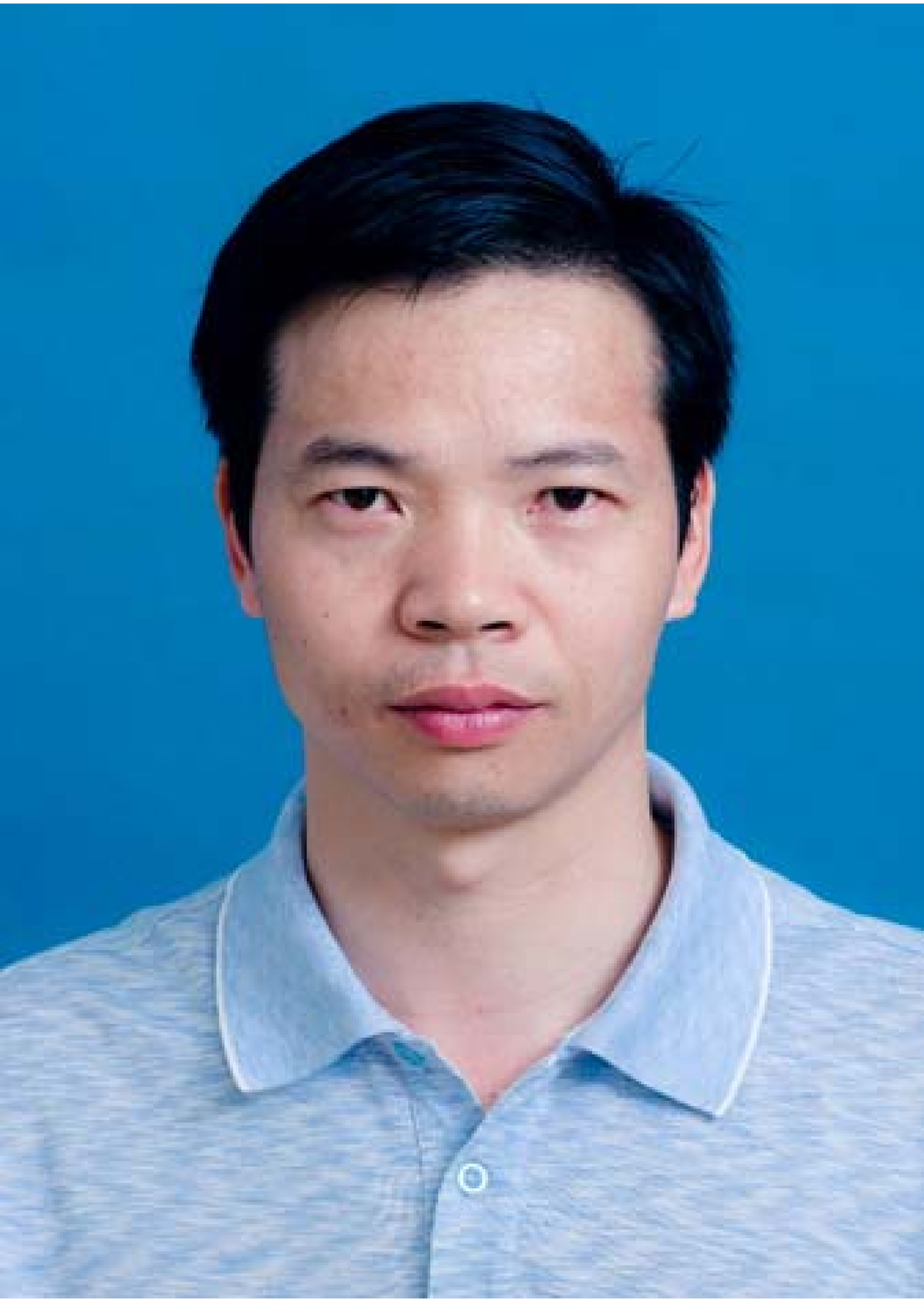}}]
{Kenli Li} received the Ph.D. degree in computer science from Huazhong University of Science and Technology, China, in 2003.
He was a visiting scholar at University of Illinois at Urbana-Champaign from 2004 to 2005.
He is currently a full professor of computer science and technology at Hunan University
and director of National Supercomputing Center in Changsha.
His major research areas include parallel computing, high-performance computing, grid and cloud computing.
He has published more than 180 research papers in international conferences and journals, such as
{\em IEEE-TC}, {\em IEEE-TPDS}, {\em IEEE-TSP}, {\em JPDC}, {\em ICPP}, {\em CCGrid}.
He is an outstanding member of CCF. He is a senior member of the IEEE and serves on the editorial board of {\em IEEE Transactions on Computers}.
\end{IEEEbiography}

\begin{IEEEbiography}
[{\includegraphics[width=1in, height=1.25in, clip, keepaspectratio]{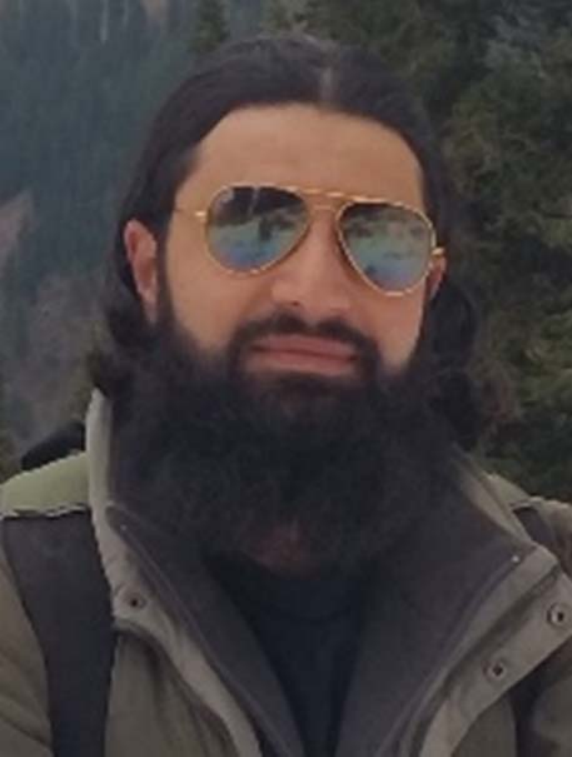}}]
{Kashif Bilal} received his PhD from North Dakota State University USA. He is currently a post-doctoral researcher at Qatar University, Qatar. His research interests include cloud computing, energy efficient high speed networks, and robustness.Kashif is awarded CoE Student Researcher of the year 2014 based on his research contributions during his doctoral studies at North Dakota State University.
\end{IEEEbiography}

\begin{IEEEbiography}
[{\includegraphics[width=1in, height=1.25in, clip, keepaspectratio]{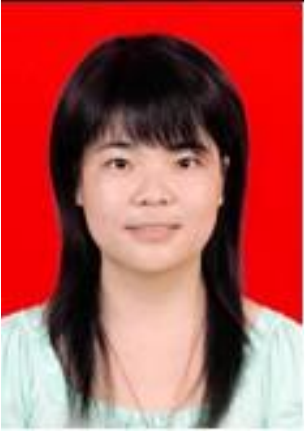}}]
{Xu Zhou} received her Ph.D. degree from the Department of Information Science and Engineering, Hunan University, in 2016. She is currently a postdoctoral in the Department of Information Science and Engineering, Hunan University, Changsha, China. Her research interests include parallel computing and data management.
\end{IEEEbiography}

\begin{IEEEbiography}
[{\includegraphics[width=1in, height=1.25in, clip, keepaspectratio]{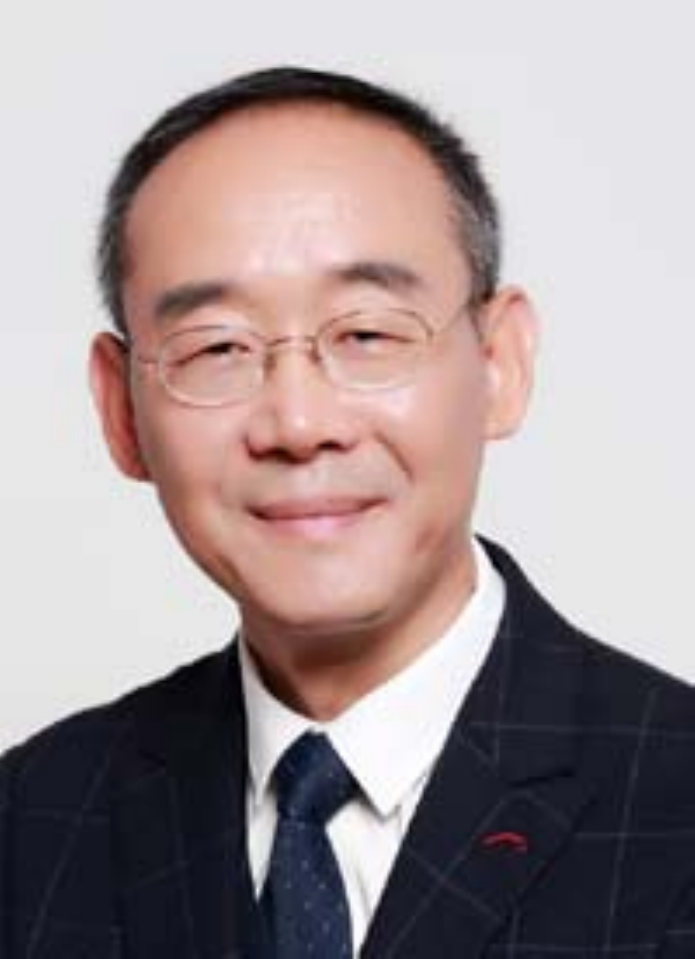}}]
{Keqin Li} is a SUNY Distinguished Professor of computer science in the State University of New York. His current research interests include parallel computing and high-performance computing, distributed computing, energy-efficient computing and communication, heterogeneous computing systems, cloud computing, big data computing, CPU-GPU hybrid and cooperative computing, multi-core computing, storage and file systems,
wireless communication networks, sensor networks, peer-to-peer file sharing systems, mobile computing, service computing, Internet of things and cyber-physical systems. He has
published over 590 journal articles, book chapters, and refereed conference papers, and
has received several best paper awards. He is currently serving or has served on the editorial boards of IEEE Transactions on Parallel and Distributed Systems, IEEE Transactions
on Computers, IEEE Transactions on Cloud Computing, IEEE Transactions on Services
Computing, and IEEE Transactions on Sustainable Computing. He is an IEEE Fellow.
\end{IEEEbiography}

\begin{IEEEbiography}
[{\includegraphics[width=1in, height=1.25in, clip, keepaspectratio]{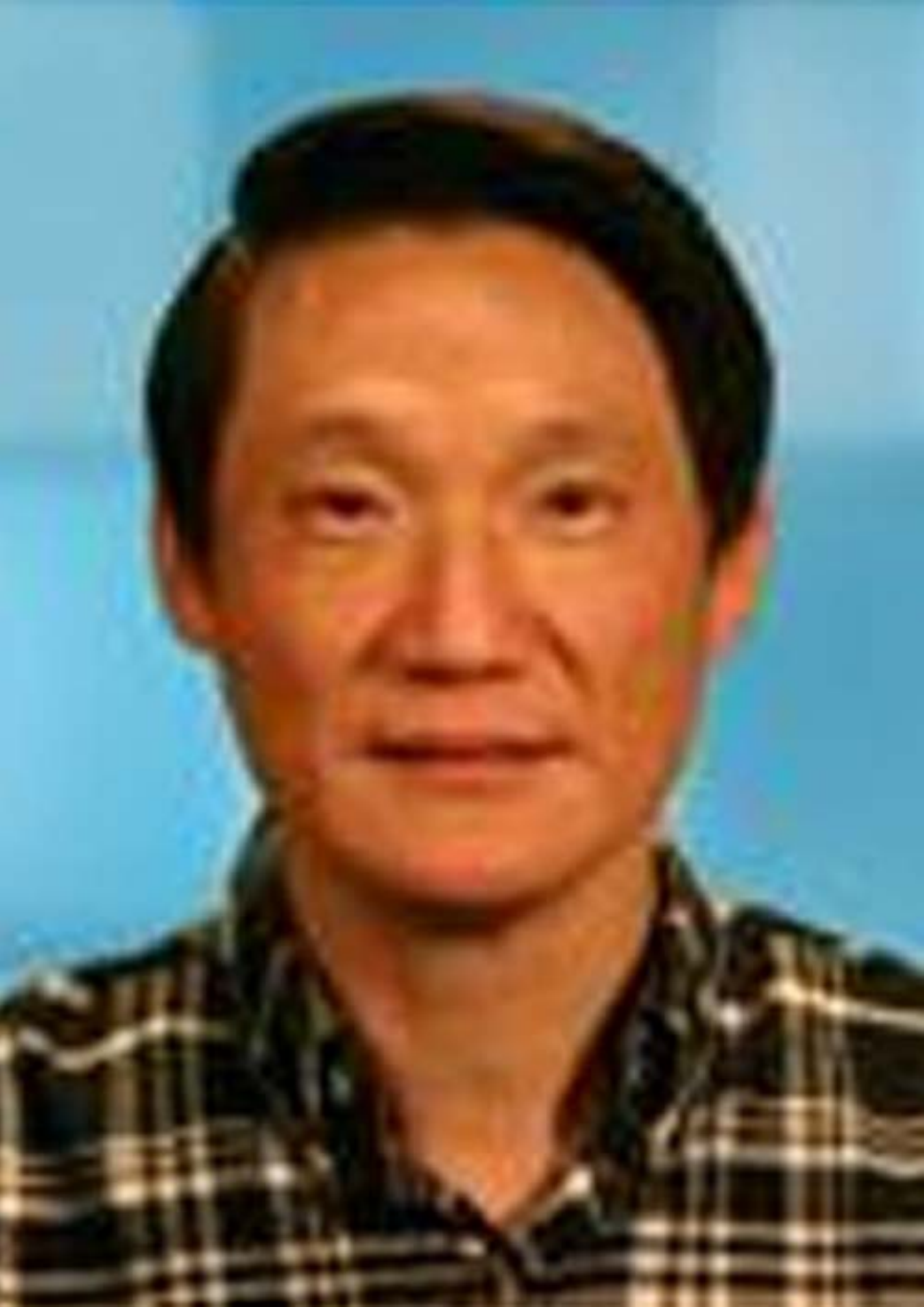}}]
{Philip S. Yu} received the B.S. Degree in E.E. from National Taiwan University, the M.S. and Ph.D. degrees in E.E. from Stanford University, and the M.B.A. degree from New York University.
He is a Distinguished Professor in Computer Science at the University of Illinois at Chicago and also holds the Wexler Chair in Information Technology.
Before joining UIC, Dr. Yu was with IBM, where he was manager of the Software Tools and Techniques department at the Watson Research Center.
His research interest is on big data, including data mining, data stream, database and privacy.
He has published more than 1,000 papers in refereed journals and conferences. He holds or has applied for more than 300 US patents.
Dr. Yu is a Fellow of the ACM and the IEEE.
He is the Editor-in-Chief of {\em ACM Transactions on Knowledge Discovery from Data}.
Dr. Yu is the recipient of ACM SIGKDD 2016 Innovation Award for his influential research and scientific contributions on mining, fusion and anonymization of big data, the IEEE Computer Society¡¯s 2013 Technical Achievement Award for ``pioneering and fundamentally innovative contributions to the scalable indexing, querying, searching, mining and anonymization of big data'', and the Research Contributions Award from IEEE Intl. Conference on Data Mining (ICDM) in 2003 for his pioneering contributions to the field of data mining.
He also received the ICDM 2013 10-year Highest-Impact Paper Award, and the EDBT Test of Time Award (2014).
\end{IEEEbiography}

\end{document}